\documentclass[sigconf, nonacm]{acmart}
\usepackage[british]{babel}
\usepackage{hyphenat}

\usepackage{graphicx}
\usepackage{balance}
\usepackage{xcolor}
\usepackage{subfig}
\usepackage{verbatim}
\usepackage{caption}
\usepackage{multirow}
\usepackage{algorithm}
\usepackage{algpseudocode}
\usepackage{comment}
\usepackage{makecell}
\usepackage{nicematrix,enumitem,booktabs}
\usepackage{pifont}
\usepackage{bbding}

\usepackage{colorprofiles}

\usepackage{float}
\floatstyle{plaintop}
\restylefloat{table}

\newtheorem{definition}{Definition}

\algnewcommand{\LineComment}[1]{\State \(\triangleright\) #1}
\providecommand{\tabularnewline}{\\}

\begin{document}
\fancyhead{}

\title{Discovering Leitmotifs in Multidimensional Time Series}

\author{Patrick Sch\"afer}
\affiliation{
  \institution{Humboldt-Universit\"at zu Berlin}
  \city{Berlin}
  \country{Germany}
}
\email{patrick.schaefer@hu-berlin.de}

\author{Ulf Leser}
\affiliation{
  \institution{Humboldt-Universit\"at zu Berlin}
  \city{Berlin}
  \country{Germany}
}
\email{leser@informatik.hu-berlin.de}

\sloppy
\begin{abstract}

A leitmotif is a recurring theme in literature, movies or music that carries symbolic significance for the piece it is contained in. When this piece can be represented as a multi-dimensional time series (MDTS), such as acoustic or visual observations, finding a leitmotif is equivalent to the pattern discovery problem, which is an unsupervised and complex problem in time series analytics. Compared to the univariate case, it carries additional complexity because patterns typically do not occur in all dimensions but only in a few - which are, however, unknown and must be detected by the method itself. In this paper, we present the novel, efficient and highly effective leitmotif discovery algorithm LAMA for MDTS. LAMA rests on two core principals: (a) a leitmotif manifests solely given a yet unknown number of sub-dimensions - neither too few, nor too many, and (b) the set of sub-dimensions are not independent from the best pattern found therein, necessitating both problems to be approached in a joint manner. In contrast to most previous methods, LAMA tackles both problems jointly - instead of independently selecting dimensions (or leitmotifs) and finding the best leitmotifs (or dimensions). Our experimental evaluation on a novel ground-truth annotated benchmark of 14 distinct real-life data sets shows that LAMA, when compared to four state-of-the-art baselines, shows superior performance in detecting meaningful patterns without increased computational complexity.
\end{abstract}

\keywords{Multidimensional, Sub-Dimensional, Multivariate, Time Series, Motif, Motif Set, Leitmotif}

\sloppy
\maketitle


\section{Introduction}

\begin{figure}[t]
    \centering
	\includegraphics[width=1.00\columnwidth]{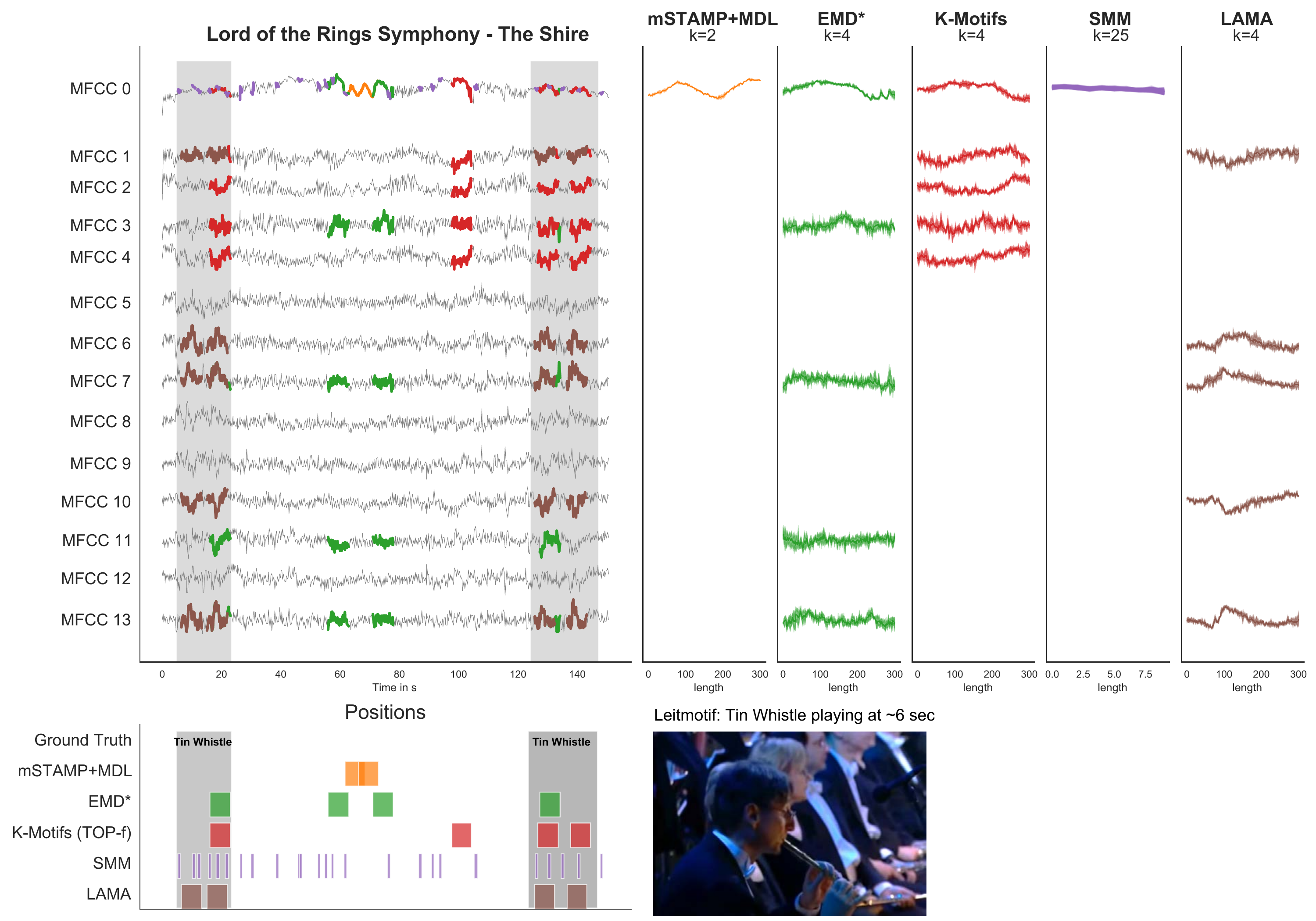}
    \Description[LOTR Use Case]{Leitmotifs found in LOTR Use Case.}
	\caption{
	\emph{The Shire} theme played by the \emph{Lord of the Rings symphonic orchestra}. The suite opens and ends with the Hobbit leitmotif, which is played on a solo tin whistle. Position of leitmotif found by LAMA (brown) compared to the ground truth (gray/bottom) and four competitor approaches (orange, green, red and purple). (Image best viewed in color)
	\label{fig:the-shire}
	}
\end{figure}

A \emph{leitmotif} (\emph{leading motif}) is a recurring theme or motif that carries symbolic significance in various forms of art, particularly literature, movies, and music. The distinct feature of any leitmotif is that humans associate them to meaning, which enhances narrative cohesion and establishes emotional connections with the audience. The use of (leit)motifs thus eases perception, interpretation, and identification with the underlying narrative~\cite{bribitzer2015understanding}. A genre that often uses leitmotifs are soundtracks, for instance in the compositions of Hans Zimmer or Howard Shore. Figure~\ref{fig:the-shire} shows a suite from \emph{The Shire} with 14 channels arranged by Howard Shore for Lord of the Rings. The suite opens and ends with the Hobbits' leitmotif, which is played by a solo tin whistle\footnote{"[Howard] Shore often arranges the Shire leitmotif for tin whistle or recorder cast in a simple homophonic texture, which can approximate the rustic countryside and carefree ways of the Hobbits."~\cite{rone2018scoring}}, and manifests in a distinct pattern in several, but not all channels of the piece. However, the discovery of leitmotifs is not limited to soundtracks. We collected $14$ real datasets from 6 domains and annotated leitmotifs.

Finding leitmotifs automatically is an unsupervised problem of high complexity, as patterns have an unknown length, occur an unknown number of times, and in an unknown number of dimensions. Furthermore, the different occurrences of a pattern are not identical but only similar to each other, where the degree of similarity may vary between the different dimensions. A specific problem is that in real-life data many dimensions often record only noise. Noise often is similar to itself and thus quickly leads to the detection of meaningless patters. Finding the sub-dimensions that contain the meaningful pattern makes the problem distinct from the univariate case~\cite{torkamani2017survey}. It adds considerable complexity, as there exist $d$ over $f$ possibilities for a fixed number of dimensions $1 \leq f \leq d$, which is $\binom{d}{f} \in \mathcal{O}\left(d^{min\left(d-f, f\right)}\right)$.

This computational burden makes brute force enumeration of possible solutions infeasible and led to the development of heuristics. To date, three strategies have been proposed. Firstly, EMD~\cite{tanaka2005discovery} pre-process the MDTS with dimensionality reduction methods, such as principal component analysis (PCA), followed by a univariate pattern finding method. Thus, it first selects (or constructs) the subset of dimensions and then finds best patterns therein. Secondly, ~\citet{minnen2007detecting} and ~\citet{yeh2017matrix} describe methods that compute patterns in each dimension separately, then sort dimensions by pattern distance, and finally find a cut-off for determining the best $f$ dimensions. Thirdly,~\citet{vahdatpour2009toward} and \citet{gao2019discovering} also first seeks patterns separately in each dimension, and then determine the subset of dimensions in a bottom-up manner by picking dimensions with correlating pattern offsets.
Our experiments (Section~\ref{sec:experiments}) with real data show that these approaches are highly noise-sensitive and often produce semantically irrelevant patterns.

In this work, we present the novel \emph{LAtent leitMotif discovery Algorithm} (LAMA), which is designed to overcome the aforementioned challenges. LAMA is the first approach to perform dimension selection and motif mining in a joint fashion. Specifically, it seeks the set of $k$ occurrences of a motif of length $l$ that has minimal pairwise distances in a latent $f$-dimensional subspace, where $l$ and $k$ can either be user-specified or learned automatically from the data. It rests on two core principles. Firstly, leitmotifs emerge from a yet unknown subset of dimensions — which must be neither too few nor too many. Secondly, the set of sub-dimensions are not independent from the best pattern found therein, which implies that both problems should be approached in a joint manner. LAMA meticulously selects each subsequence as a potential leitmotif candidate and constructs a candidate set by considering the subsequence along with its k-nearest neighbors (k-NN). In a joint learning process, LAMA determines the most suitable dimensions for each candidate by assessing the k-NN distance within each dimension. This involves sorting the k-NN distances in ascending order and selecting the dimensions with the lowest distances. To find the best leitmotif, LAMA minimizes the pairwise distance among the constituent subsequences within all candidate sets of size $k$. This ensures the identification of the optimal Leitmotif across various sub-dimensional candidate sets.

Figure~\ref{fig:the-shire} shows the results of five core approaches LAMA, EMD*~\cite{tanaka2005discovery}\footnote{We re-implemented the relevant parts of EMD, as the source code is not available.}, mSTAMP~\cite{yeh2017matrix} using MDL for channel selection, SMM~\cite{poccia2021smm}, and K-Motifs~\cite{lonardi2002finding}, with aggregated distances over the first $5$ channels (same as for other competitors), details in Chapter~\ref{sec:related–work}. We use the same motif length $l\approx5s$ for all approaches to avoid bias. mSTAMP~\cite{yeh2017matrix} identifies two occurrences of a seemingly straight line located in the loudness channel (MFCC 0), equal to a quiet section of the song. We reason this to be the fade out of one of the instruments within the Hobbits' leitmotif. EMD* finds a motif set with only $2$ out of the $4$ Hobbits' leitmotif correctly identƒified, and $2$ unrelated subsequences. Notably, even the similarity between subsequences of the same channels is low - see channels $7$ or $11$ resulting from the plot. K-Motifs use no channel explicit selection strategy. Using the first five channels it fails at identifying all occurrences with $3$ out of $4$ occurrences found, and one seemingly unrelated sub-sequence. SMM does not provide a way to define the motif length. Consequently, even with tuning, it detects 25 occurrences of what seems to be a short straight line in the loudness channel. Our LAMA is the only method to correctly identify $4$ occurrences within the leitmotif. Other than EMD*, LAMA's occurrences show high pairwise similarity, too.

In summary, our paper makes the following contributions:
\begin{enumerate}
    \item We introduce LAtent leitMotif discovery Algorithm (LAMA), an algorithm to uncover multi-dimensional leitmotifs. Notable, LAMA distinguishes itself as the pioneer by jointly tackling the dual challenge of selection sub-dimensions and discovering motif sets - unlike SotA, which prioritize one aspect over the other (Section~\ref{sec:subdimensional-motif-sets}).
    \item A critical step in LAMA is the computation of a distance matrix of TS windows across dimensions. We show that a naive implementation would lead to excessive memory and present a new algorithm that reduces memory consumption by up to three orders of magnitude.
    \item 
    We describe innovative algorithms for automatically determining two of the three hyper-parameters of LAMA, which makes its application considerably easier compared to competitors, for which users have to define manually up to 7 user parameters.
    \item We present the first annotated benchmark of 14 real-life MDTS, ranging from the distinctive punching motion in boxing, the fundamental steps of a dance routine, to the leitmotif of Darth Vader in Star Wars.
\end{enumerate}

We assess the performance of LAMA through a novel benchmark comprising 14 real-life manually labelled MDTS datasets. Our evaluation contrasts LAMA against four state-of-the-art competitors: EMD*, mSTAMP, SMM, and K-Motifs, examining both the accuracy of pattern discovery against annotated ground truth and the computational runtime. We analyze the impact of four distance measures and noise on accuracy. Additionally, we demonstrate LAMA's superior leitmotif detection quality through three compelling case studies spanning diverse domains, including the Star Wars Imperial March (soundtracks), Rolling Stones (pop music), and Boxing Routines (motion data). We find that LAMA finds considerably more meaningful patterns (Leitmotifs) in each of these data sets. 

The rest of the paper is organized as follows: In Section~\ref{sec:background} we present the background and definitions. Section~\ref{sec:sub-dimensonal-motif-sets} presents our Leitmotif approach. Section~\ref{sec:experiments} shows the utility of our method throughout various case studies. Finally, Section~\ref{sec:related–work} gives an overview of related work for multidimensional motif set discovery

\section{Background}\label{sec:background}

In this section, we first formally define time series (TS) and the z-normalized Euclidean distance (z-ED), which we (like all prior work) will use throughout this work. We then introduce the definition of Leitmotifs for sub-dimension motif set discovery.

\begin{definition}
\emph{Multivariate Time Series (MTS)}: A multivariate time series $T=\left(t_1, \ldots t_n \right) \in \mathbb{R}^{(d \times n)}$ is a list of $n$ vectors with each $t_i \in \mathbb{R}^d$ being a vector of $d$ dimensions (sometimes referred to as channels). We denote the $i$-th observation of the $k$-th dimension by the scalar $t^{(k)}_{i}\in \mathbb{R}$, and the $k$-th dimension by $T^{(k)}\in \mathbb{R}^n$
\end{definition}

If every point in $t_{i} \in T$ in the TS represents a single value ($t_{i} \in \mathbb{R}$), the series is \emph{univariate}. If each point represents the observations of multiple variables (e.g., temperature, humidity, pressure, etc.) we call the TS \emph{multivariate} or \emph{multidimensional}. 

\begin{definition}
\emph{Multivariate Subseq.}: A subsequence $S_{i;l}$ of a MTS $T = (t_1, \dots, t_n) \in \mathbb{R}^{(d \times n)}$, with $1 \leq i \leq n$ and $1 \leq i+l \leq n$, is a MTS of length $l$ consisting of the $l$ contiguous vectors, each with $d$ dimensions, from $T$ starting at offset i: $S_{i;l} =(t_i,t_{i+1},...,t_{i+l-1}) \in \mathbb{R}^{(d \times l)}$.
\end{definition}

Works in motif discovery (MD) typically exclude overlapping subsequences, as their distance is naturally low.

\begin{definition}
\emph{Overlapping subsequences (Trivial Match)}: Two subsequences $S_{i;l}$ and $S_{j;l}$ of length $l$ of the same TS $T$ overlap iff $\lceil i-l\cdot\alpha \rceil \le j \le \lceil i+l\cdot\alpha \rceil$ where $\alpha \in [0,1]$.
\end{definition}
By default, we set $\alpha=1.0$, which excludes any overlapping subsequences sharing more than $1$ value. However, this parameter can be adjusted through our API.

The z-normalized Euclidean distance (z-ED) is the prevalent distance measure for MD. It can be formulated as a dot-product. This facilitates rapid computation of the distance matrix among all pairwise subsequences sampled from a TS using MASS~\cite{zhong2024mass}, with complexity $\mathcal{O}(n^2)$, regardless of subsequence length $l$. Internally, MASS calculates a rolling dot product in constant time. This enables the straightforward computation of z-ED, Euclidean distance (ED), cosine similarity (CS), or Complexity Invariant Distance (CID)~\cite{batista2011complexity}. Our implementation supports all four distance measures. For brevity, we will focus on z-ED. We first define z-ED within a single channel, then extend it to the multivariate case.

\begin{definition}
\emph{z-normalized Euclidean distance (z-ED)}:
Given two univariate subsequences $A=(a_1, \ldots, a_l)$ with mean $\mu_A$ and standard-deviation $\sigma_A$ and $B=(b_1, \ldots, b_l)$ with $\mu_B$ and $\sigma_B$, both of length $l$, and dot-product $A \cdot B$ their (squared) z-ED is defined as:
\begin{equation}\label{eq:z-ED}
\textit{z-ED}^2(A, B)=2l \left( \frac{A \cdot B - l \mu_A \mu_B}{l \sigma_A \sigma_B}\right)
\end{equation}
\end{definition}

In the case of multivariate subsequences, we conceptually compute the distance between A and B in each dimension, separately, and sum up the result.

\begin{definition}
\emph{Multivariate z-ED}:
Given two multivariate subsequences $A=(a_1, \ldots, a_l)$, with $k$-th dimension $A^{(k)} \in \mathbb{R}^{(n)}$, and $B=(b_1, \ldots, b_l)$, with $k$-th dimension $B^{(k)} \in \mathbb{R}^{(n)}$ respectively, both of length $l$ and dimension $d$, their multivariate distance is defined as:
\begin{equation}
\textit{dist}(A, B)=\sum_{k=1}^{d}\textit{z-ED}^2\left(A^{(k)},B^{(k)}\right) \label{eq:m_ed}
\end{equation}
\end{definition}

Instead of using all $d$ dimensions, we often want to compute the distance between two multidimensional subsequences using only a subset of dimensions $X$, with $|X|=f \leq d$.

\begin{definition}
\emph{Sub-dimensional Distance}:
Given two multivariate subsequences $A=(a_1, \ldots, a_l)$ and $B=(b_1, \ldots, b_l)$, both of length $l$ and $a_i, b_i \in \mathbb{R}^{d}$, their sub-dimensional distance using only the subset of dimensions $X$ is defined as:
\begin{equation}
\textit{dist}^{(X)}(A, B)=\sum_{k \in \mathcal{X}}\textit{dist}\left(A^{(k)},B^{(k)}\right) \label{eq:sub-dimensional-distance}
\end{equation}
where $X$ is the subset of dimensions, and $|X|=f$
\end{definition}

To derive sub-dimensional patterns, the initial stage involves identifying the \emph{pertinent sub-dimensions} that collectively \emph{minimize the overall distance}. Given the number of dimensions $f$ as input, the corresponding optimization problem is to find the closest subset $X^{*}$ of dimensions, with $|X^{*}|=f \leq d$, for which the above distance between two subsequences $A$ and $B$ in Eq.~\ref{eq:sub-dimensional-distance} is minimal:
$$
X^{*}= \min_X \textit{dist}^{(X)}(A, B)
$$

To avoid high similarity between overlapping subsequences extracted from a TS, we define $e$-matches as:

\begin{definition}
\emph{e-match}: Two subsequences $A$ and $B$ of $T$ are called $e$-matching iff (a) $\textit{dist}^{(X)}(A, B) \leq e \in \mathbb{R}$ and (b) they are \emph{non-overlapping}. A set $S$ of subsequences of $T$ is called $e$-matching, iff all subsequences in $S$ are pairwise $e$-matching.
\end{definition}

We next define the \emph{extent} of a motif set.

\begin{definition}\label{def:extent}
\emph{Extent}: Consider a MTS $T$, a set $S$ of subsequences of $T$ of length $l$, and sub-dimensions $X$. The \emph{extent} of $S$ is the maximal pairwise distance of elements from $S$:
$$ e = extent(S, X) = \max_{(A, B) \in S \times S} \textit{dist}^{(X)}(A, B)$$
\end{definition}

$k$-Motiflets~\cite{schafer2022motiflets}, are the set $S^*$ of exactly $k$ occurrences of a motif of length $l$, whose maximal pairwise distance is minimal.
\begin{equation}\label{eq:kmotiflet}
\begin{split}
S^{*} & = \min_{S, |S|=k} extent(S)
\end{split}
\end{equation}

A heuristic solution~\cite{schafer2022motiflets} for finding $k$-Motiflets involves a $(k-1)$-nearest neighbor search around each subsequence of a TS, and choosing the set $S^*$ with minimal extent, where $S*$ includes the query and its $k-1$-NN.

Next, we introduce our novel concept, the \emph{Leitmotif}, along with its distinctive minimization problem:
\begin{definition}\label{def:leitmotif}
\emph{Top Leitmotif}: Given a multidimensional TS $T$ of dimensionality $d$, and (i) parameters cardinality $k \in \mathbb{N}$, (ii) motif length $l$, and (iii) number of sub-dimensions $f \leq d$, the top Leitmotif of $T$ is the set $S$ with $|S|=k$ subsequences of $T$ of length $l$ for which the following holds: All elements of $S$ are pairwise $e$-matching, with $e=extent(S)$, and there exists no set $S'$ with $extent(S') < extent(S)$ also fulfilling these constraints.
\end{definition}

\begin{definition}\label{def:leitmotif_optimiation}
The Leitmotif optimization problem is then defined as finding the most similar (closest) $|X^{*}|=f$ dimensions and $|S^{*}|=k$ subsequences for which holds:
\begin{equation} 
\begin{split}
X^{*}, S^{*} & = \min_{X, S} extent(S, X)
\end{split}
\end{equation}
\end{definition}

The Leitmotif optimization problem naturally extends the $k$-Motiflet optimization problem, which was originally defined for uni-dimensional motif set discovery~\cite{schafer2022motiflets}. When using a single dimensional TS as input, the two definitions are equivalent. As finding $k$-Motiflet exactly is likely NP-hard~\cite{schafer2022motiflets}, heuristics are required.

\section{Leitmotif Discovery}\label{sec:sub-dimensonal-motif-sets}



We first address the challenges in finding sub-dimensional leitmotifs (Section~\ref{sec:challenges}). We then introduce our novel Leitmotif algorithm \textsc{LAMA} for discovering sub-dimensional motif sets in multivariate TS, providing an overview in Section~\ref{sec:overview} and our heuristic solution in Section~\ref{sec:multivariate-motif-sets}. Section~\ref{sec:subdimensional-motif-sets} discusses issues with sub-dimensional motif discovery and our solution. Algorithms to learn the input parameters $k$ and $l$ are presented in Section~\ref{sec:variable-length-motif-sets}, followed by computational complexity analysis (Section~\ref{sec:complexity}). Finally, we discuss scalable Leitmotif discovery (Section~\ref{sec:scalable-motif-sets}).

\subsection{Finding Sub-Dimensional Leitmotifs}~\label{sec:challenges}

We've distilled two fundamental principles guiding leitmotif discovery. Firstly, leitmotifs emerge from an elusive, yet-to-be-defined subset of dimensions, striking a balance between not being too sparse nor too abundant. Secondly, the sub-dimensions set must be individually uncovered for each leitmotif candidate. For instance, consider the solo tin flute consistently playing within the same channel subset across all instances in our motivational example. LAMA addresses both challenges simultaneously, avoiding the sequential approach of dimension learning followed by motif discovery. Given the computational infeasibility of exhaustively enumerating all sub-dimensions, we use heuristics. These heuristics prioritize either channel selection or motif discovery first. We'll explore the advantages and limitations of the state-of-the-art approach~\cite{yeh2017matrix} for set discovery in Section~\ref{sec:subdimensional-motif-sets}.



\subsection{Overview of LAMA}~\label{sec:overview}

Mathematically, we defined a \emph{Leitmotif} as a collection of $k$ approximate occurrences of a subsequence with a length of $l$, with minimal extent in sub-dimensional space. In a first pre-processing phase, the length $l$ and the number of repetitions $k$ of a motif are learned from sensible input ranges (see Section~\ref{sec:variable-length-motif-sets}.
\begin{figure}[t]
    \centering
	\includegraphics[width=1.00\columnwidth]{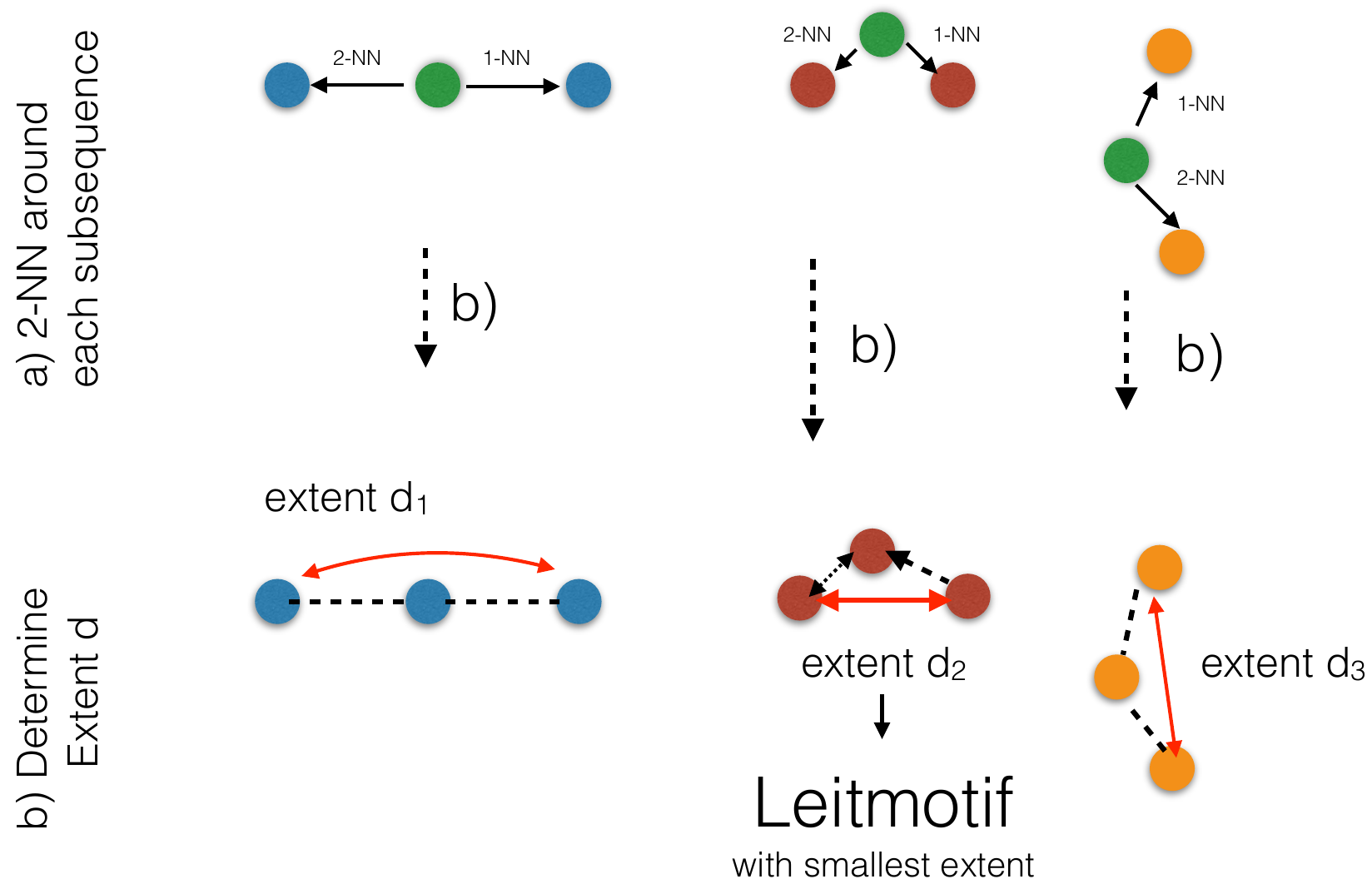}
    \Description[Leitmotif Discovery]{Exemplary LAMA Leitmotif Discovery.}
	\caption{
	Depicted are three sets (blue, red, orange), centered around a query (green point) and its $2$-NN. Leitmotif discovery involves two steps: (a) $2$-NN search around each query subsequence, and (b) determine the extent of each set (red arrow), i.e. $d_1, d_2, d_3$. Finally, the top Leitmotif with smallest extent $d_2$ is returned.
	\label{fig:approximate_motiflet_discovery}
	}
\end{figure}
Figure~\ref{fig:approximate_motiflet_discovery} illustrates the core idea behind LAMA for computing a leitmotif with $3$ occurrences. LAMA iteratively perform two steps for each subsequence $q$ (in green): (a) search for the $(k-1)$-NN of $q$, and (b) determine their (sub-dimensional) pairwise extent of the candidate set. Finally, the set with minimal (sub-dimensional) extent is returned (in red at bottom).

\begin{figure}[t]
    \centering
	\includegraphics[width=1.0\columnwidth]{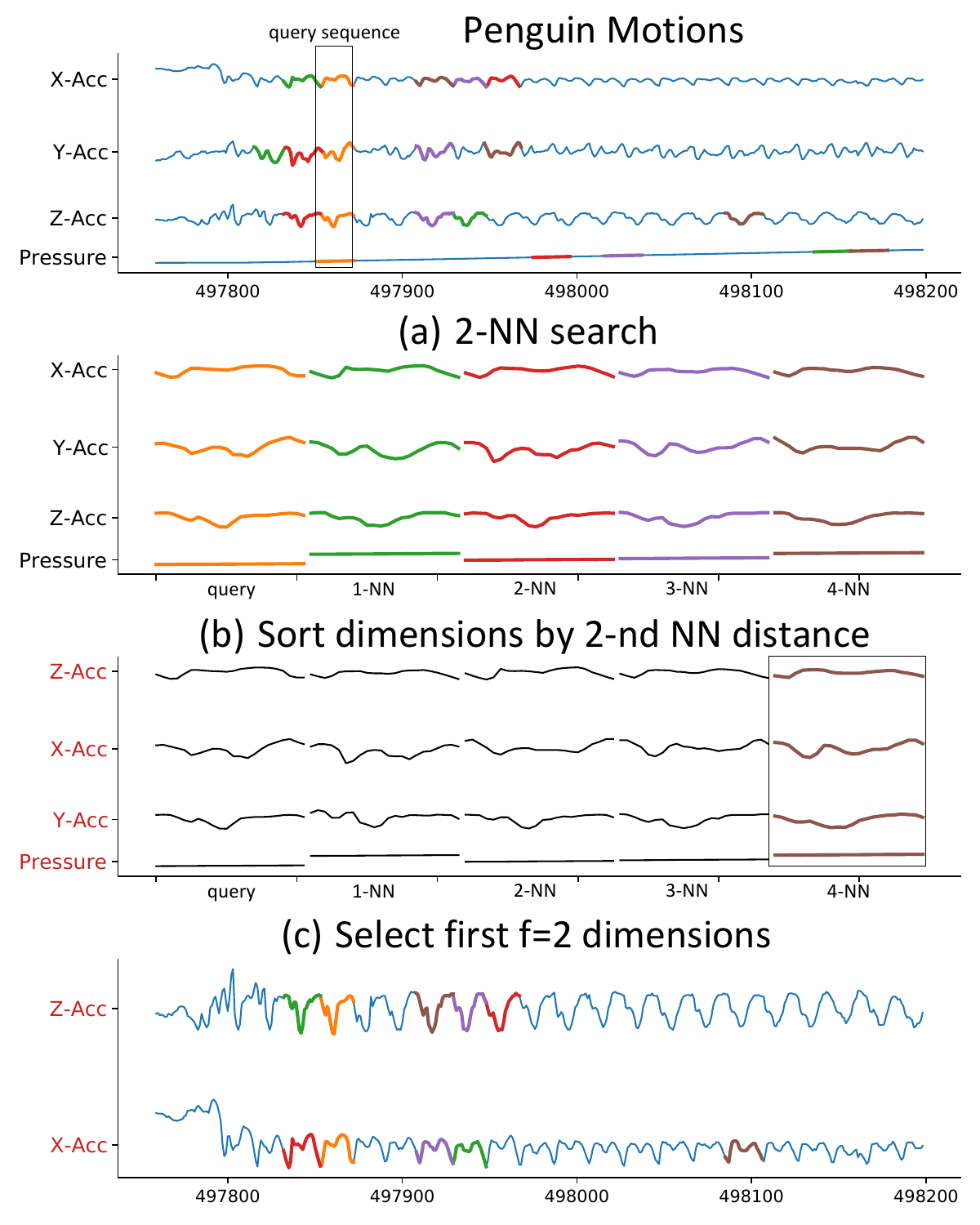}
    \Description[Dimensionality Reduction Workflow]{Dimensionality Reduction Workflow used in LAMA.}
	\caption{
	Dimensionality Reduction Workflow used in LAMA. Top: a query sequence is selected. (a) For each of query sequence, a k-NN search is performed along each dimension. (b) The dimensions get sorted by the distance of the k-th NN. (c) The first $f$ dimensions are kept for this sequence.\label{fig:workflow}
	}
\end{figure}

The workflow for jointly detecting sub-dimensions with leitmotif discovery in LAMA is illustrated in Figure~\ref{fig:workflow}, and will be detailed in the following sections. It shows $4$ dimensions of a penguin diving from~\cite{zhu2017matrix}. For each multivariate query subsequence (Figure~\ref{fig:workflow} TOP in orange), LAMA performs a 2-NN search (Figure~\ref{fig:workflow} (a)) in each dimension. It next sorts the dimensions by the distance of their 2-th NN (Figure~\ref{fig:workflow} (b)), which results in the order of Z-Acc, X-Acc, Y-Acc, Pressure. Finally, LAMA selects the best $f$ dimensions (i.e. f=2), based on lowest distance, from the TS. We may then take either of the k-NNs in dimension Z-Acc or X-Acc as Leitmotif candidate. We decided to take the first dimensions k-NNs as candidate (see Section~\ref{sec:theoretical-bounds} for the rationale).

To find the best Leitmotif, LAMA minimizes the extent over all sub-dimensional candidate sets of size $k$. For each candidate set, an optimal set of sub-dimensions is determined.


\subsection{\emph{LAtent leitMotif discovery Algorithm} (LAMA)}~\label{sec:multivariate-motif-sets}
\begin{algorithm}[t]
    \small
	\caption{Compute Leitmotifs using LAMA}\label{alg:k-leitmotifs}
	\begin{algorithmic}[1]
		\Procedure{RUN\_LAMA}{$T$, $D$, $f$}
            \State $(l,k) \gets \textsc{learn\_parameters}(T, D)$
            \State \Return $\text{LAMA}(T, D, f)$
        \EndProcedure
		\Procedure{LAMA}{$T$, $D$, $l$, $k$, $f$}
        \State $d \gets \text{dimensions}(T)$
        \State $n \gets \text{len(T)} - l + 1$
        \State $(leitmotif, e) \gets (\{\}, inf)$ \Comment{bsf of extent}
        \For{$i \in [1, \dots, n] $}
            \LineComment{\emph{Stage 1: Determine sub-dimensions}}
            \State $kNNs \gets \text{matrix of size } d \times k$
            \For{$dim \in [1, \dots, d] $}
                \State $kNNs[dim] \gets$ \textsc{non\_trivial\_argkNN}$(D[dim, i], k)$
            \EndFor
            \LineComment{\emph{Stage 2: Check for leitmotif}}
            \State $f\_dims, candidate \gets$ \textsc{select\_f\_dimensions}$(D, kNNs, i, f)$
            \State $dist_{knn} \gets$ $\sum_{dim \in f\_dims} D[dim, i, kNNs[dim, k]]$
            \If{$dist_{knn}< e$}
                \State $dist$ $\gets$ \textsc{pairwise\_extent}$(D, f\_dims, candidate, e)$
                \If{$dist < e$}
                    \State ($leitmotif, e) \gets (candidate, dist)$
                \EndIf
            \EndIf
        \EndFor
		\State \Return $(leitmotif, e)$
		\EndProcedure
	\end{algorithmic}
\end{algorithm}
LAMA continuously undergoes the joint tasks for each leitmotif candidate: (a) it identifies the set of relevant sub-dimensions, and (b) the extent is minimized to find the best (TOP) leitmotif. Algorithm~\ref{alg:k-leitmotifs} illustrates our approach for discovering the Leitmotif. The \textsc{LAMA} procedure takes as input the TS $T$, the pairwise distance matrix $D$, of dimensionality $d \times (n-m+1) \times (n-m+1)$, computed separately along each dimension of the TS, and a hyper-parameter for the number of sub-dimensions $f$ to use. The two parameters cardinality $k \in \mathbb{N}$, and motif length $l \in \mathbb{N}$ are learned from the TS (Section~\ref{sec:variable-length-motif-sets}). Using $D$, we can compute full or sub-dimensional pairwise distances between any pair of multivariate subsequences using Eq.~\ref{eq:m_ed}. To make use of admissible pruning, we use the observation that the $k$-NN distance of a subsequence can be used to derive a lower and an upper bound on the extent of the candidate set, which effectively reduces the number of extent computations. Specifically, the extent must be at most $2$ times the distance of the k-NN distance:
$$e \leq dist_{kNN} \leq 2 \cdot e \label{eq:lower_bound}$$
Intuitively speaking, the extent of a motif set must be between the radius and diameter, when two subsequences are at opposite bounds, of a hypersphere centered around the query subsequence (Figure~\ref{fig:approximate_motiflet_discovery}).

\textsc{LAMA} applies admissible pruning to reduce the number of candidate sets using the lower bound on the extent, using k-NN distances $dist_{knn}$, and the current best-so-far extent $e$. The algorithm processes all subsequences (line~9) sequentially in a greedy fashion. It approximates the extent of the optimal set of subsequences using the $(k-1)$-NNs of a query to form a candidate set for the Leitmotif.

The NN have to be non-overlapping to each other to avoid any trivial matches in extent computation. In $\textsc{non\_trivial\_argkNN}()$ (line~13) we order the subsequences by their distance to the $i$-th subsequence and return the closest $(k-1)$ non-trivial, i.e. non overlapping, neighbours in each dimension. Using these NNs we select the $f$ best dimensions using $\textsc{select\_f\_dimensions()}$, line~16.  We will discuss this in the following Section~\ref{sec:subdimensional-motif-sets}. For the time being, assume we selected all dimensions to be used from $D$. The distance of the $(k-1)$-th NN is used for admissible pruning (line~18ff). The extend of a candidate set has to be checked, if and only if $dist_{kNN}$ is lower than the current best-so-far extent $e$. If successful, we determine the pairwise extent of this set (line~19), again using the best $f$ dimensions. To speed up computations, we apply admissible pruning in $\textsc{pairwise\_extent}(\_, \_, \_, e)$, too, by stopping the computation once any pairwise distance exceeds $e$. If the overall extent $dist$ is smaller than the best-so-far $e$, we update the Leitmotif (lines~21). Once all $n$ candidate offsets have been processed, we return the Leitmotif and its extent $e$.

\subsection{Sub-dimensional Leitmotifs}~\label{sec:subdimensional-motif-sets}

Selecting the optimal sub-dimensions to minimize the extent of the Leitmotif is a challenging task. Theoretically, the process of choosing $f$ out of $d$ dimensions results in $d$ over $f$  possibilities, which is in $ \binom{d}{f} \in \mathcal{O}\left(d^{min\left(d-f, f\right)}\right)$ for fixed $f$. This computational burden makes direct computation practically infeasible.

\subsubsection{Independent Dimensionality-Selection using Pairs}

Fortunately, a log-linear solution, leveraging sorting techniques, has been proposed for addressing this challenge, particularly when minimizing the distance between pairs of subsequences in~\cite{minnen2007detecting}. This approach has found success in \emph{pair motif} discovery~\cite{yeh2017matrix}. However, its efficacy diminishes when applied to the task of identifying \emph{motif sets}, and we will delve into the reasons behind its limitations. Subsequently, we propose an extension, enabling the discovery of motif sets (Leitmotifs) in log-linear time in dimensions $d$.

The extent is defined as the maximum over all pairs of distances within the motif set $S$, and we need to find those sub-dimensions $X$, that minimize the overall extent to find the optimal set $S^{*}$. The objective is given by:
\begin{equation}
\min_{S, X} extent(S, X)
\end{equation}

This objective becomes minimal, by expanding the definition of the extent, and find the best sub-dimensions for each pair of sub-sequences, such that:
\begin{equation} \label{eq:independent}
\begin{split}
S^{*} & = \min_{S, X} extent(S, X)   \\
      & = \min_{S} \left( \max_{(A, B) \in S \times S} \left( \min_{X} \textit{dist}^{(X)}(A, B) \right) \right)
\end{split}
\end{equation}

This formulation entails first selecting a motif set $S$, followed by selecting sub-dimensions $X$ with minimal distance for each pair of subsequences, to compute the maximum extent of the set. Minimizing yields the motif set $S^{*}$  with overall lowest extent.

Furthermore, this formulation allows for a log-linear efficient implementation. For a multi-dimensional TS, with time and channel axis, finding the k-NN involves an ascend sort by distance along time within each channel, and choosing the first $k$-th entries. This is followed by sorting along the channels in ascending order, and always choosing the first $f$ channels with smallest channel distance~\cite{minnen2007detecting, yeh2017matrix}. This works well for pair motifs, i.e. the 1-NN~\cite{yeh2017matrix}. Yet, it tends to find less intuitive motif sets, as it independently optimizes motif discovery and channel selection, as discussed next.


\begin{figure}[t]
    \centering
    \includegraphics[width=1\columnwidth]{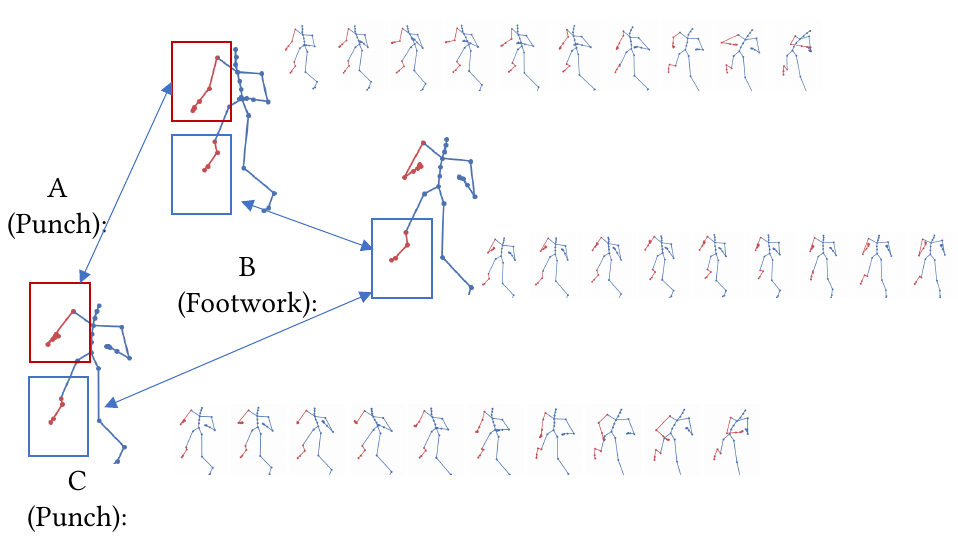}    \Description[Boxing Routine]{Similarity of Motifs in Boxing Routine.}
	\caption{
	The figure shows three boxing routines A, B, C (each from left to right), which are unintuitively considered similar when independently optimizing/selecting dimensions between pairs (compare Eq.~\ref{eq:independent}). While A and C show a punching motion, in B the actor simply turns his body around his own axis making small steps. (A, B) and (B, C) show similarity in steps, while (A, C) is similar in the punching motion. (Image best viewed in color)
	\label{fig:boxing}
	}
\end{figure}

Consider the example illustrated in Figure~\ref{fig:boxing}. The example shows a boxing motion capture. Employing sub-dimensional leitmotif search, we find three subsequences denoted as A, B, C. Notably, subsequences $A$ and $B$ exhibit similarity in the punching motion (red). $(B, C)$ and  $(A, B)$ share similarity in the foot-steps (blue). Despite minimal pairwise similarity in the selected dimensions, the overall motif set encompasses three distinct motions. This is a result of independent motif set discovery and channel selection.

\subsubsection{Joint Dimensionality Selection using Sets}
To address the aforementioned issue, we reformulate the optimization target. Our objective is to identify sub-dimensions $X^{*}$, with $|X^{*}|=f$, and subsequences $S^{*}$, with $|S^{*}|=k$, that exhibit highest similarity among all subsequences of the same set:
\begin{equation}\label{eq:dependent}
X^{*}, S^{*} = \min_{X} \left(\min_{S} \left(\max_{(A, B) \in S \times S} \textit{dist}^{(X)}(A, B) \right) \right)
\end{equation}

How are the two equations, Eq.\ref{eq:independent} and Eq.\ref{eq:dependent}, different? Let's first consider a toy example with two sets of distances to illustrate the effect of swapping min and max: Set $S_1: \{1, 3, 5\}$ and Set $S_2: \{2, 4, 6\}$. First, compute the max of each set: $\max(S_1) = 5$ and $\max(S_2) = 6$. The min of these is $5$. Next, compute the min of each set: $\min(S_1) = 1$ and $\min(S_2) = 2$. The max of these is 2. By moving the max operation outside, we get a higher total distance, thus \emph{extent} of the leitmotif.

Conceptually, when moving the minimization target sub-dimensions $X$ outside the distance computations, as in Eq.\ref{eq:dependent}, we jointly learn a fixed set of sub-dimensions that lead to an optimal candidate leitmotif. This approach ensures that motif candidates share a common set of dimensions, such as the same feet or hand movements, to be considered similar. In contrast, Eq.\ref{eq:independent} identifies independent sub-dimensions for each pair of subsequences. This results in subsequences to be considered similar if any part of the body performs a similar movement between any two subsequences, even if these parts differ between other pairs of subsequences. Overall, this results in a less cohesive leitmotif.

\subsubsection{Solution}

Regrettably, there is currently no optimal and efficient solution available to address this joint minimization problem, necessitating the enumeration of all $d$ over $f$ solutions to identify the optimal one. An effective approximation is to select the best $f$ dimensions based on minimizing the distance between the query and the $(k-1)$-NN, which we will present next.

\subsubsection{Algorithm}\label{sec:sub-dimensional-algorithm}

Algorithm~\ref{alg:subdim-k-leitmotifs} illustrates our approach. It takes as input the pairwise distance matrix D, of dimensionality $d \times (n-m+1) \times (n-m+1)$, computed separately along each dimension of the TS, the kNNs along each dimension, with dimensionality $d \times (n-m+1) \times k$, the position $i$ of the query sequence in the TS,  and the number of dimensions $f$ to choose from $d$.

It first extracts the distance for the ($k-1$)-th NN in each dimension (lines 5--7) prior to sorting these distances (line~9) and choosing the first $f$. From these $f$ dimensions, the set of ($k-1$)-NN from the first dimension is used as Leitmotif candidate (lines 11-12).

\begin{algorithm}[t]
    \small
	\caption{Sub-Dimension Selection (f out of d)}\label{alg:subdim-k-leitmotifs}
	\begin{algorithmic}[1]
		\Procedure{ select\_f\_dimensions}{$D$, $kNNs$, $i$, $f$}
            \State $d \gets \text{dimensions}(kNNs)$
            \LineComment{1. Get $k^{th}$-NN distance by dimension}
            \State $distkNN \gets \textit{Array of size dims}$
            \For{$dim \in [1, \dots, \text{d}] $} \Comment{(k-1)-th NN by dimension}
                \State $distkNN[dim] \gets$ $D[dim, i, kNNs[dim]]$
            \EndFor
            \LineComment{2. Sort and take best $f$ dimensions}
            \State $f\_dims \gets$ \textsc{first}$_f($ \textsc{argsort}$(distkNN))$
            \LineComment{3. Use first (lowest distance) dimension as candidate}
            \State $candidate$ $\gets$ $kNNs[f\_dims[0]]$

    		\State \Return $(f\_dims, candidate)$
		\EndProcedure
	\end{algorithmic}
\end{algorithm}

\subsection{Learning Length and Set Size from the Data}~\label{sec:variable-length-motif-sets}

We next present methods to learn suitable values for the Leitmotif length $l$ and set size $k$ to ease the discovery of motifs can be found without domain knowledge. These are based on an analysis of the extent function for different $k$, as introduced in~\cite{schafer2022motiflets}:
\begin{definition}
\emph{Extent Function (EF)}: Assume a fixed length $l$ and a TS T. Let $S_k$ be the top $k$-Motiflet with length $l$ of $T$. Then, the \emph{extent function} $EF$ for $T$ is defined as $EF(k) = extent(S_k)$.
\end{definition}

Using the EF lead to the following two ideas~\cite{schafer2022motiflets}:
\begin{figure}[t]
    \centering
	\includegraphics[width=1.0\columnwidth]{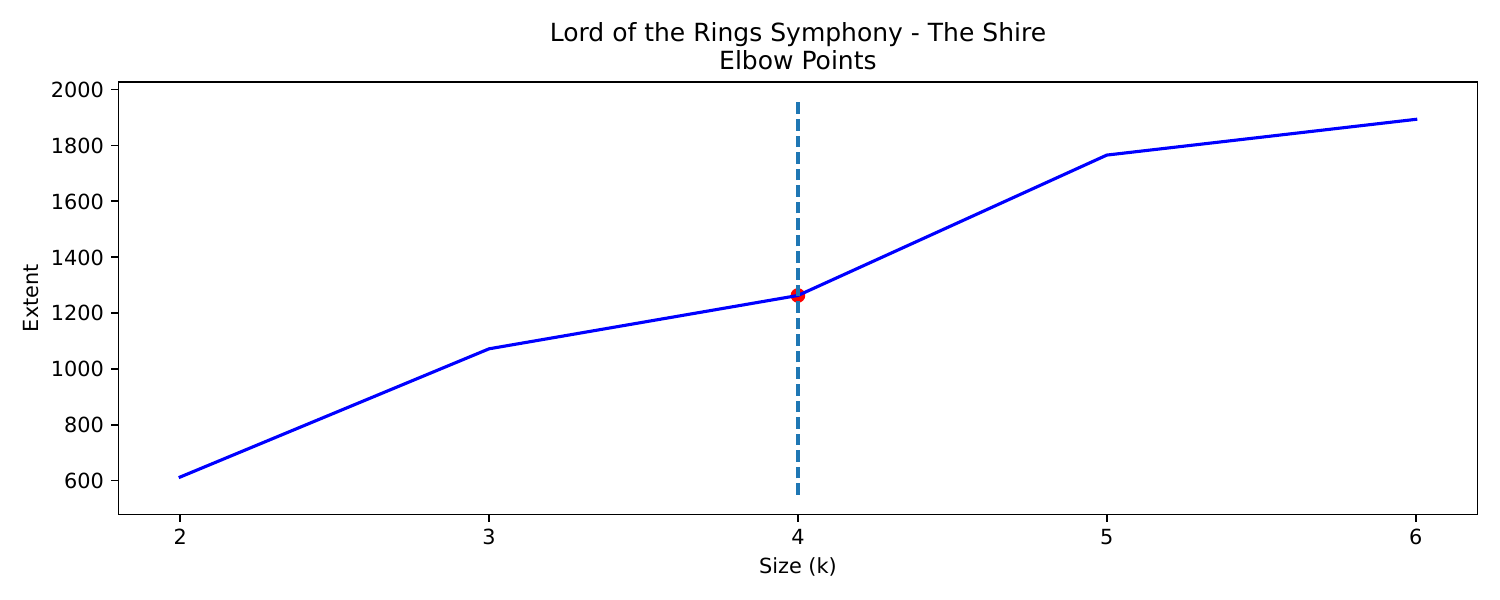}
    \Description[Elbows]{Elbows for LOTR Use Case.}
	\caption{
	The \emph{EF} is a function of the cardinality of motif set to its extent. Elbow points represent large changes in similarity of the found motif, indicative of a concept change.
	\label{fig:elbow_method}
	}
\end{figure}

\begin{figure}[t]
    \centering
	\includegraphics[width=1.0\columnwidth]{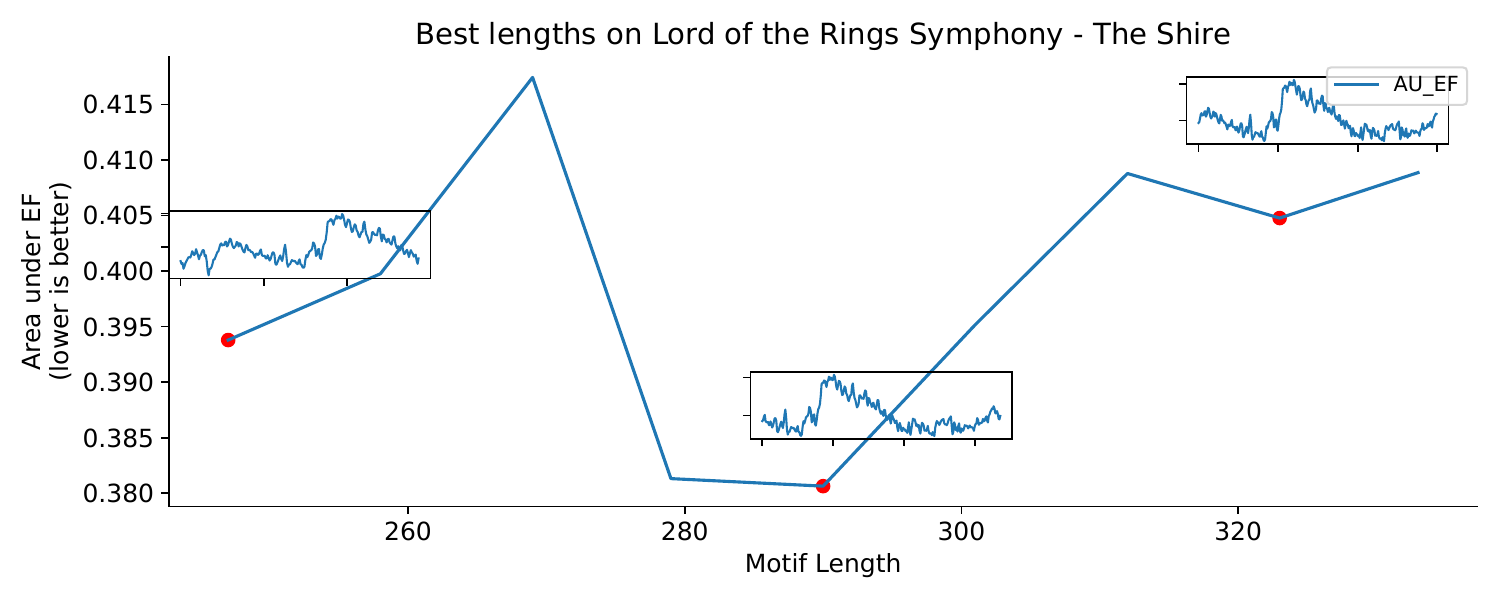}
    \Description[Area under Elbow Function]{Area under the elbow function on LOTR Use Case.}
	\caption{
	\label{fig:window_length_selection}
    Area under the elbow function points to interesting lengths at local minima.
	}
\end{figure}

\begin{enumerate}[label=\textbf{\arabic*.}, left=0pt, labelsep=1em, itemsep=0.5em]
\item Elbow points in the \emph{EF} indicate changes of motifs. The last value of $k$ before the elbow indicates a maximal motif, and is considered a meaningful value for $k$. Figure~\ref{fig:elbow_method} shows the introductory \emph{The Shire} use case. An elbow is detected at $k=4$ equal to the four repetitions of the leitmotif.
\item Long flat stretches of the EFs indicate a high number of occurrences of a motif, but depend on the motif length $l$. Accordingly, window lengths $l$ leading to long flat stretches are considered as particularly meaningful, which can be measured using the area under the EF (AU-EF).
\end{enumerate}

Variable-length motif discovery involves two-fold: computing motifs along a range of lengths, and ranking the motifs found to reduce redundancy. For variable-length Leitmotifs, we compute the motif along a range of motif lengths $l in [l_{min}, l_{max}]$, and report those lengths, which show a local minimum in the AU-EF plot. These represent lengths, where the elbow-function is flat, and thus contain a motif set with the largest possible number of repeats. The concept is illustrated in Figure~\ref{fig:window_length_selection} for \emph{The Shire} dataset. 

\subsection{Computational Complexity}~\label{sec:complexity}

We will next analyse the complexity of algorithm~\ref{alg:k-leitmotifs}. It consists the building blocks: (a) extracting $k$-NN along each dimension, (b) computing the extent, and (c) selecting sub-dimensions.

\emph{Regarding (a)}: Extracting $k$ nearest neighbors in $d$ dimensions over all subsequences takes $\mathcal{O}(d \cdot n^2)$ using linear time selection (Quick-Select+Median of Medians) on each subsequence.\\ 
\emph{Regarding (b)}: Selecting $f$ out of $d$ sub-dimensions involves sorting the distances of the $k-1$-th NN along $d$ dimensions, which is performed $\mathcal{O}(n)$ times. Using MergeSort, this totals $\mathcal{O}\left(n\cdot \left(d \log d \right)\right)$.\\
\emph{Regarding (c)}: The computation of the pairwise extent takes $\mathcal{O}(d \cdot k^2 \cdot n)$ in the worst case, if admissible pruning is not successful. The best case is in $\mathcal{O}(d \cdot k^2)$, if the first candidate leads to admissible pruning of all other candidates.

Overall, the complexity of the algorithm using $d$ dimensions is in $\mathcal{O}\left(d \cdot \left(n^2 + n \log d + k^2\right)\right)$ in the best case,  and $\mathcal{O} \left( d \cdot \left(\textbf{k} \cdot n^2 + n \log d + k^2\right)\right)$ in the worst case.

\subsubsection{Theoretical Bounds}\label{sec:theoretical-bounds}

The extent must be at most $2$ times the distance of the k-NN distance. Figure~\ref{fig:approximate_motiflet_discovery} left illustrates the worst case, when two subsequences are at opposite sides of the hyper-sphere. Thus, the $k$-NN distance serves as an upper bound for the extent: $e \leq dist_{kNN} \leq 2 \cdot e$

By sorting distances of the dimensions for each subsequences in ascending order, and choosing the sub-dimensions based on the k-NN (Algorithm~\ref{alg:subdim-k-leitmotifs} line~9), we minimize this upper bound (the $2\cdot e$ part). I.e. choosing a dimension with a larger k-NN distance yields a potentially larger upper bound.  There is no guarantee though, as choosing a different subset of dimensions may still yield a smaller extent $e$ of the Leitmotif. A same argument holds for choosing a different set of candidates other than from the first dimension (Algorithm~\ref{alg:subdim-k-leitmotifs} line 11).
Overall, this approach is a $2$-approximation to choosing the optimal subset of dimensions. Future work might include algorithms to provide better bounds.

\subsection{Scalable Motif Discovery}~\label{sec:scalable-motif-sets}

One inherent limitation of the LAMA algorithm is scalability in terms of memory, which is quadratic in the size of the input. A (squared) distance matrix of a 30k (50k) TS requires roughly 3 GB (9 GB) of RAM. Thus, main memory quickly becomes a bottleneck for MD on large TS. A non-sufficient naive approach to reduce the memory, would be to only store the distance of each subsequence to its $k-1$-NN, aka, $k$-dimensional distance profile. However, to compute the extent of a candidate set, we need to compute the distance between all pairs of subsequences (compare Eq.~\ref{def:extent}), i.e. the distance between each NN.


The core idea is to compute the distance matrix twice in a row wise manner, always keeping only the distances in that row. When computing distances row-by-row, we only have access to the distance of the current subsequence $i$ to its $k$-NN. Within these $k$-NNs there are pairs for which we have not seen any distance yet, i.e. when $j > i \textit{ with } j \in \textit{ k-NNs(i)}$. In the first pass, we compute the distance matrix row-by-row. For each row, we memorize the required pairwise distances from the candidate set. These offsets can point to preceding or succeeding rows of the distance matrix, which are not accessible. In the second pass, we iterate the matrix in a row-by-row manner again, and keep only those distances that were memorized in the first pass.
This optimization reduces memory complexity by 2 to 3 orders of magnitude, at the cost of doubling the distance matrix computation complexity. To reduce runtime complexity, we case study finding leitmotifs in a TS segmentation in Section~\ref{sec:streaming}.

\section{Experimental Evaluation}\label{sec:experiments}

Our experimental evaluation is two-fold. Firstly, in Section~\ref{sec:quantative} we compare our LAMA against SotA in a \textbf{quantitative} analysis using precision and recall on a benchmark dataset with  manually labelled ground truth. Secondly, in Section~\ref{sec:qualitative} we compare methods in terms of the \textbf{quality} of results on multiple case studies.

\subsection{First Annotated Leitmotif Benchmark Set}

No public benchmark exists for sub-dimensional motif discovery. Table~\ref{tab:datasets} provides an overview of the 14 datasets used in our evaluation. The Boxing, Ice Ice Baby, and Vanilla Ice datasets were used in previous studies~\cite{yeh2017matrix, kamgar2019matrix} without ground truth annotation. The other datasets are original use cases. We selected datasets where the leitmotif can be easily verified, such as audio or video, where refrains or common motions are recognizable by humans but challenging for algorithms. We manually inferred the ground truth positions for each dataset and provided gold standard parameters, including the size $k$ of the motif set, motif length, and the number of relevant dimensions $f$.

\emph{\textbf{Four Motion Captures}} Recordings originate from the CMU archive~\cite{de2009guide}. Each consists of $31$ joints, subdivided into x, y, z coordinates, with a total of $93$ dimensions, at a frame rate of $120$ frames/second.
    \emph{Charleston} contains a recording of a female subject dancing 20's Charleston, a partner dance originated in the golden 20's. The leitmotif is the basic footstep with swinging feed and arm movements, which is performed three times with a duration of $~1.3s$.
    \emph{Boxing} captures of boxing routine. The leitmotif is a punch with the right arm with 10 repetitions of $~1.6s$ each.
    \emph{Swordplay} is a recording of a series of Japanese sword strokes. Each sword movement is distinct, but the common leitmotif is the raising of the sword above the head with $6$ repetitions of $~1s$.
    \emph{Basketball} shows the dribbling of a basketball, while an actor is moving forwards. The dribbling motion is repeated five times, which $~0.42$ each.

\emph{\textbf{Two soundtracks}:} A genre that often uses leitmotifs are soundtracks. Each song has $20$ MFCC channels.
    \emph{Howard Shore} uses a large number of leitmotifs to support characters or the landscape of Middle-earth. The song \emph{The Shire} contains the Hobbits' leitmotif with four occurrences of $8-9s$ each.
    \emph{John Williams} used leitmotifs to represent people and concepts in Star Wars, i.e., one leitmotif is played in the presence of Darth Vader. This leitmotif in the \emph{Imperial March} is the trumpets playing, which is repeated $5$ times with $~8.3s$.

\emph{\textbf{Five pop songs}:} In pop songs a Leitmotifs can be a refrain, chorus, a rhythm section, or the bass line. Each song has 20 MFCC channels.
    \emph{Rolling Stones' Paint it Black} was released in 1966 on the Aftermath album. The sitar is featured prominently throughout the song, which adds a unique Middle Eastern flair. The verses of the song breakdown in a similar verse structure of B, C, B, C, B, C, D, C, B, C, E, F, where the C section is repeated most often. Its leitmotif is roughly $5s$ long and repeated twice per C section with a total of 10 times.
    \emph{Linkin Park - Numb} was released in 2003 on the Meteora album. The refrain is repeated $6$ times with $~5.5s$ each.
    \emph{Linkin Park - What I've done} was released in 2007 on the Minutes to Midnight album. The refrain is repeated $6$ times with $~10s$ each.
    \emph{Queen - Under Pressure} was performed by Queen and David Bowie, and published in 1981 on the Greatest Hits album. It is most famous for its baseline, which was repeated by \emph{Vanilla Ice - Ice Ice Baby} with a minor modification. This baseline is repeated $16$ times for Under Pressure and $20$ times for Ice Ice Baby, and $~4.2$s.

\emph{\textbf{Wildlife Recording}:} We used a wildlife recording of birds in their natural habits, provided by the Royal Society for the Protection of Birds (RSPB)~\cite{rspb2023}. It has 20 MFCC channels. Bird vocalizations have been a challenging test case for TS MD~\cite{imani2019matrix}. The common starling is a noisy bird, with the male being the primary singer. Its song comprises various melodic and mechanical sounds arranged in a ritualistic succession. The leitmotif contains repeated clicks and concludes with a burst of high-frequency sounds of $4$ repetitions of $~1.3s$ each.

\emph{\textbf{Crypto/Stocks}:}
    \emph{Bitcoin Halving Event:} This dataset includes daily closing values for seven cryptocurrencies over ten years, focusing on the price movements of Bitcoin following its halving events. Historically, Bitcoin prices rise significantly after each halving, typically over the course of a year. The leitmotif contains the $3$ halving events that affect Bitcoin and Litecoin with 1 year duration.

\emph{\textbf{Wearable Sensor Data}:}
    \emph{Physiodata:} This dataset comprises inertial and magnetic sensor data from wearable devices during physical therapy exercises~\cite{YURTMAN2014189}. It includes an exercises performed in three ways (correct, fast, low-amplitude). The motif is repeated $20$ times and $~6s$ long.

\begin{table}
\caption{Datasets used in experiments. Ground leitmotifs were manually inferred. GT refers to the number of leitmotif occurrences.\label{tab:datasets}}
\small
\begin{NiceTabular}{c|c|c|c|c}
\toprule
Use Case & Category & Length & Dim. & GT\tabularnewline
\midrule
Charleston~\cite{de2009guide} & Motion Capture & 506 & 93 & 3 \tabularnewline
Boxing~\cite{de2009guide} & Motion Capture & 4840 & 93 & 10\tabularnewline
Swordplay~\cite{de2009guide} & Motion Capture & 2251 & 93 & 6\tabularnewline
Basketball~\cite{de2009guide} & Motion Capture & 721 & 93 & 5\tabularnewline
LOTR - The Shire & Soundtrack & 6487 & 20 & 4\tabularnewline
SW - The Imperial March & Soundtrack & 8015 & 20 & 5\tabularnewline
RS - Paint it black & Pop Music & 9744 & 20 & 10\tabularnewline
Linkin Park - Numb & Pop Music & 8018 & 20 & 5\tabularnewline
Linkin P. - What I've Done & Pop Music & 8932 & 20 & 6\tabularnewline
Queen - Under Pressure & Pop Music & 9305 & 20 & 16\tabularnewline
Vanilla Ice - Ice Ice Baby & Pop Music & 11693 & 20 & 20\tabularnewline
Starling~\cite{rspb2023} & Wildlife Rec. & 2839 & 20 & 4\tabularnewline
Physiodata~\cite{YURTMAN2014189} & Wearable Sensors & 5526 & 5 & 20\tabularnewline
Bitcoin Halving & Crypto/Stocks & 3591 & 7 & 3\tabularnewline
\bottomrule

\end{NiceTabular}

\end{table}

\textbf{Competitors:} Achieving reproducibility proved to be a major challenge. Many of the approaches have roots dating back one or two decades~\cite{tanaka2005discovery,minnen2007detecting,vahdatpour2009toward}, and unfortunately, none of the codes for Multidimensional Motif Set  Discovery (MMSD) competitors have been publicly shared. We reached out to the two competitors within the last 5 years, namely TMDM~\cite{balasubramanian2016discovering} and SMM~\cite{poccia2021smm} asking for code, but only obtained the code for SMM.
We use the following approaches: (a) \emph{mSTAMP (MDL)}, a pair motif approach, which uses the Minimum Description Length (MDL) for sub-dimension selection, (b) \emph{mSTAMP} using the $f$-dimensions from ground truth annotations, (c) \emph{EMD*}~\cite{tanaka2005discovery}, an approach using PCA for channel reduction followed by univariate motif set discovery, (d) \emph{K-Motifs (TOP-f)}, a univariate method that aggregates the distances for the first $f$ channels, and (e) \emph{K-Motifs (all dims)} that aggregates distances over all channels, and (f) \emph{SMM}.

\textbf{Parameters:} MD is an unsupervised learning task. \textsc{LAMA} has $1$ hyper-parameter with the number of dimensions $f$, and two parameters ($k$ and $l$) that can be learned from the data given parameter ranges without access to any labelling. If learning was applied, the concrete ranges and learned parameters are defined within each experiment. For the experiment with ground truth available, we use the associated parameters as input to all competitors. For SMM we tested the ranges: \textsc{DeGaussianThres:} from 0 to 1, \textsc{DeLevelTime} 4, 6, 8, \textsc{DeLevelDepd:} 4, 6, 8, \textsc{DeSigmaDepd:} 0.4, 0.5, 0.6, and \textsc{r:} 5 and 10. We provide source codes and results on our \textbf{website}~\cite{LeitmotifWebPage}.

\subsection{Quantitative Comparision}\label{sec:quantative}

\begin{figure}[t]
    \centering
	\includegraphics[width=0.5\columnwidth]{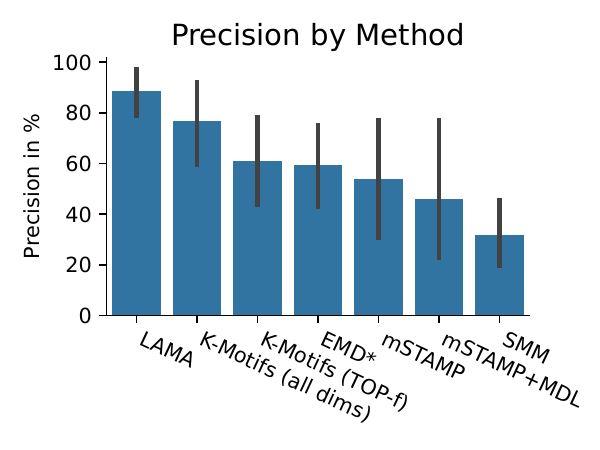}\includegraphics[width=0.5\columnwidth]{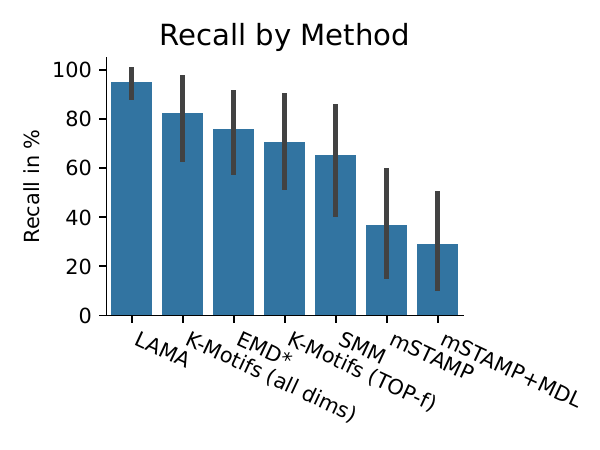}
    \Description[Precision and Recall]{Precision and Recall on Benchmark Datasets}
	\caption{Use cases with inferred semantic leitmotifs: Mean Precision (left) and Mean Recall (right) of the found motif sets by MD method. LAMA performs best.
	\label{fig:barplots}}
\end{figure}

To measure the accuracy of the different approaches, we created a benchmark of ground-truth annotated datasets (Table~\ref{tab:datasets}) by manually inspecting the data for semantic motif sets. We use the gold standard parameters as input to the methods: the size $k$ of the motif set, and the length of the motif set, or the number of relevant dimensions $f$.
We assume that a method finds a motif set if the reported motif interval overlap with a ground truth interval by at least $50\%$. Figure~\ref{fig:barplots} shows the results as (a) \emph{precision}, measured as percentage of correct prediction out of all predictions, and (b) \emph{recall}, measured as percentage of ground-truth annotations found.

\emph{LAMA} has by far the highest precision and recall, see Table~\ref{tab:results} with a margin of $12$ to $42$ percentage points for mean precision and $12.5$ to $66$ for mean recall. This underlines that jointly finding sub-dimensions and motifs is crucial for capturing leitmotifs. LAMA performs worst for the datasets \emph{Vanilla Ice - Ice Ice Baby} and \emph{Queen David Bowie - Under Pressure}, which both contain the famous bass line.
The competitors struggle to find ground truth annotations. The best competitor is the univariate method \emph{K-Motifs (all dims)}, which simply sums the distances computed over each dimension to derive a univariate signal. \emph{K-Motifs (TOP-f dims)}, which selects the TOP-f dimensions to sum up distances, has a significantly lower precision and recall. This underlines the need for choosing the ideal subset of dimensions rather than any subset. \emph{mSTAMP} in both variants shows by far the worse mean and median recall, with median recall being $0$ and $20$, as it must only return pairs of motifs, whereas the motifs range from $4$ to $20$ occurrences (Table~\ref{tab:datasets}). Mean precision is slightly better, but still the worst among competitors.
Channel selection using MDL is biased towards selecting noisy or near constant channels. \emph{EMD*} globally selects dimensions using PCA, based on explaining the variance of the whole TS. This fails to capture the semantic motif sets, which is captured in small sections of a TS.
SMM has $7$ hyper-parameters that require tuning. Despite tuning efforts, SMM consistently returned between 1 and 50 motifs per dataset, each with up to 80 locations. We selected the best-fitting motif for each dataset, which inherently provides an advantage. Nevertheless, SMM's precision remains suboptimal. Further analysis revealed that many motifs have a length of 1, making them single points rather than subsequences, and they are limited to a single dimension. Even after seeking guidance from the author, tuning the seven parameters of SMM remains the sole bottleneck.

\begin{table}
\caption{Precision and Recall by Method.\label{tab:results}}
\small
\begin{tabular}{lrrrr}
\toprule
 & \multicolumn{2}{r}{Precision} & \multicolumn{2}{r}{Recall} \\
 & mean & median & mean & median \\
Method &  &  &  &  \\
\midrule
EMD*                &  59.3 &   65.0 &  75.9 &   80.0 \\
K-Motifs (TOP-f)    &  61.1 &   70.0 &  70.8 &  \textbf{100.0} \\
K-Motifs (all dims) &  76.8 &   83.3 &  82.6 &  \textbf{100.0} \\
mSTAMP              &  53.9 &  \textbf{100.0} &  36.7 &   20.0 \\
mSTAMP+MDL          &  46.2 &    0.0 &  29.0 &    0.0 \\
SMM                 &  31.8 &   26.5 &  65.4 &   95.0 \\
LAMA                &  \textbf{88.7} &  \textbf{100.0} &  \textbf{95.1} &  \textbf{100.0} \\
\bottomrule
\end{tabular}
\end{table}

\begin{figure}[t]
    \centering
	\includegraphics[width=0.5\columnwidth]{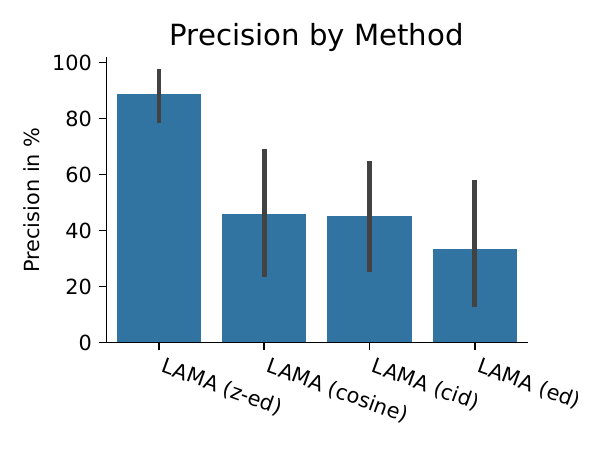}\includegraphics[width=0.5\columnwidth]{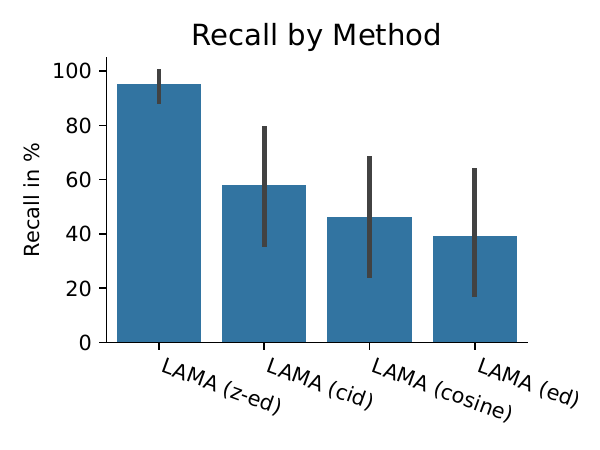}
    \Description[Precision and Recall]{Precision and Recall on Benchmark Datasets}
	\caption{Experiment with four different distance measures on benchmark dataset. z-ED performs best.
	\label{fig:barplots_dists}}
\end{figure}

\emph{Distance Measures:}
We conducted experiments using the four implemented distance measures and LAMA. Figure~\ref{fig:barplots_dists} presents the results. Overall, the z-normalized ED (z-ED) achieved the highest precision and recall on our benchmark, followed by cosine distance (CD), complexity invariant distance (CID), and standard ED.
The results indicate that ED and CD are highly sensitive to low-variance segments of the signal, such as the resting phase in the physio-data. CID performs slightly better as it compensates for low variance. One exception is \emph{Vanilla Ice}, where ED outperforms z-ED by effectively capturing the baseline.

\begin{figure}[t]
    \centering
	\includegraphics[width=1.0\columnwidth]{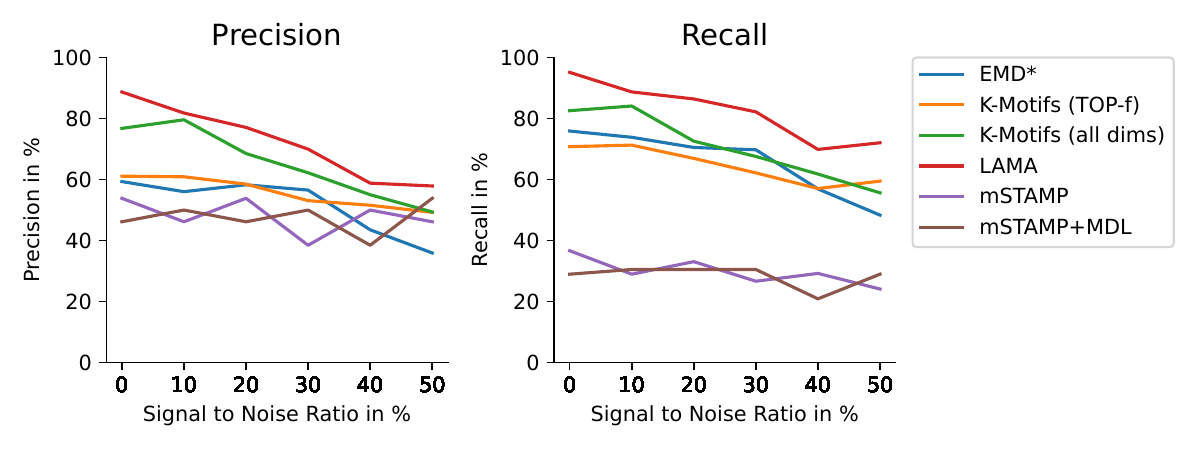}
    \Description[Precision and Recall]{Precision and Recall on Benchmark Datasets}
	\caption{Experiment with increasing noise levels.
	\label{fig:lineplot_noise}}
\end{figure}

\emph{Influence of Noise:}
Finally, we conducted experiments with increasing noise levels by adding Gaussian noise to the series, up to 50\% of the dataset's standard deviation (Figure~\ref{fig:lineplot_noise}). The addition of noise makes it progressively harder to distinguish between the leitmotif and low variance parts of the series. Consequently, the precision and recall of almost all methods decrease as the noise level increases, complicating the detection. Despite this, LAMA consistently achieves the highest scores across all noise levels. As mSTAMP initially identified low-noise sections of a signal irrelevant to the leitmotif, this makes it the least affected by noise.

\subsection{Qualitative Comparision}\label{sec:qualitative}

In this section we discuss the quality of the discovered motif sets.
The purpose is to compare methods not only by precision and recall of found motifs as in the previous section, but also to consider whether these motifs are actually meaningful, i.e., correspond to important concepts in the TS. For each experiment we first ran LAMA to learn optimal value for the length of the leitmotif $l$. As other methods offer no method for learning input parameters, we then use this length $l$, the number of occurrences $k$, and the number of channels $f$ associated with the ground truth of each dataset as input to all methods for a fair comparison. SMM returns up to $50$ motifs, with the best matching motifs consisting of sets of dozens of points that are barely visible to the eye (compare Figure~\ref{fig:the-shire}). Thus, we chose to display SMM results only on our webpage.

\paragraph{\textbf{Motion Capture}}

In this experiment we case study the analysis of motion capture data by identifying characteristic behaviour. Motion data is challenging to due its high number of dimensions, with $93$ dimensions, for $3$1 joints each with $x, y, z$ axis.

\begin{figure}[t]
    \centering
	\includegraphics[width=1.0\columnwidth]{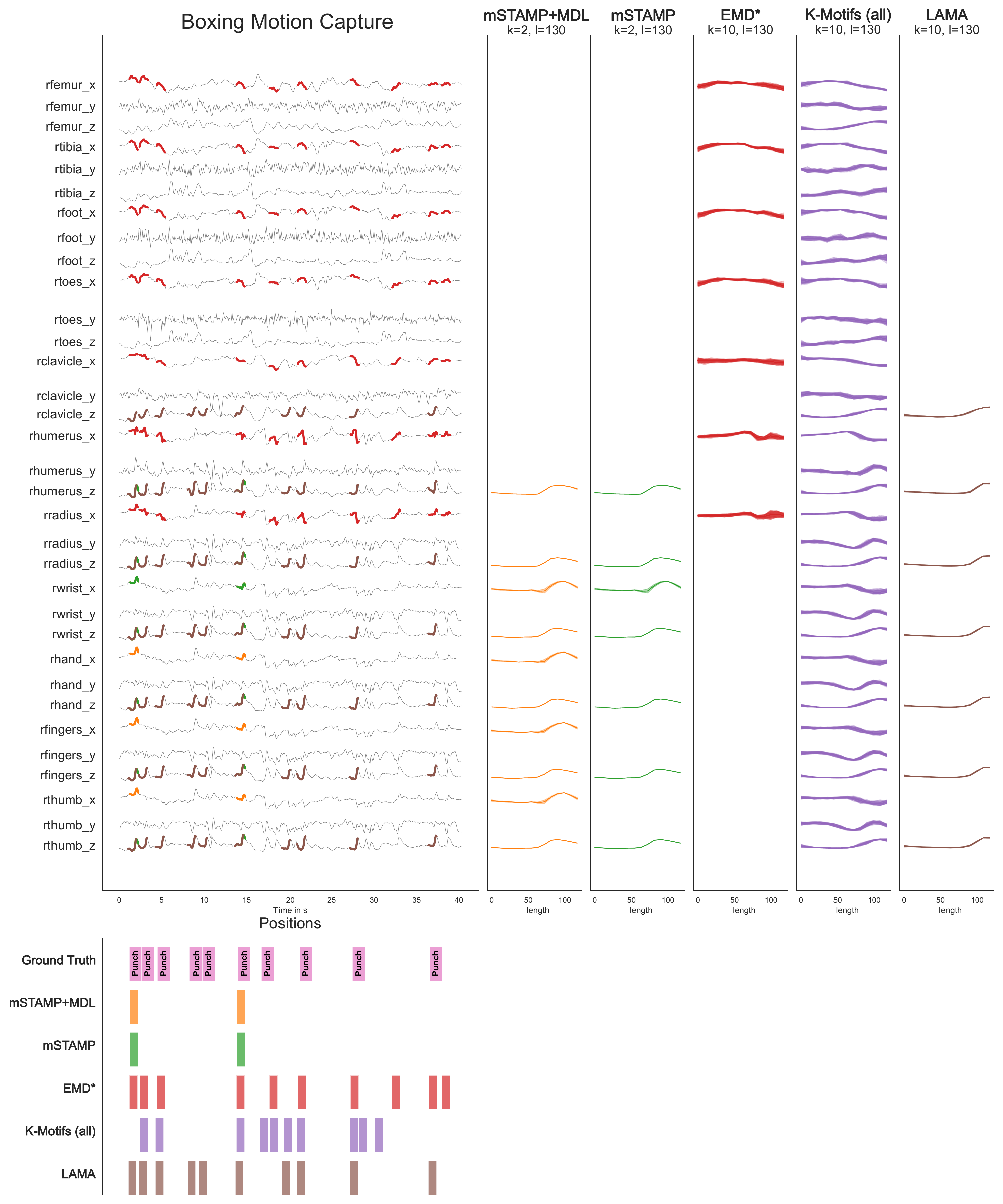}
    \Description[Boxing Motion]{Leitmotifs found in Boxing Motions Case Study}
	\caption{
	Punching motions found in a motion capture of a \emph{Boxing routine}. (Image best viewed in color)
	\label{fig:boxing_new}
	}
\end{figure}

Figure~\ref{fig:boxing_new} shows a TS of a boxing routine. As part of the routine, the actor performs multiple other motions. The characteristic motion (leitmotif) is a punch/swing with the right hand. The punch is repeated 10 times with a duration of roughly $1.6s$ each. We only used the joints of the right body halve. mSTAMP and LAMA mostly agree in important dimensions for punching, which are located at the right clavicle, humerus, radius, wrist, hand, fingers, and thumb. Other than that EMD* identifies mostly joints at the right leg, which seems unusual for punching, and represents the triple steps performed prior to a punch. As such, LAMA is the only method to identify all punching motions. mSTAMP, in both variants, identifies two punches. EMD* fails to identify all punches, correlating to identifying feed movement. K-Motifs similarly fails.

\paragraph{\textbf{Pop Music - Paint it Black}}

\begin{figure}[t]
    \centering
	\includegraphics[width=1.0\columnwidth]{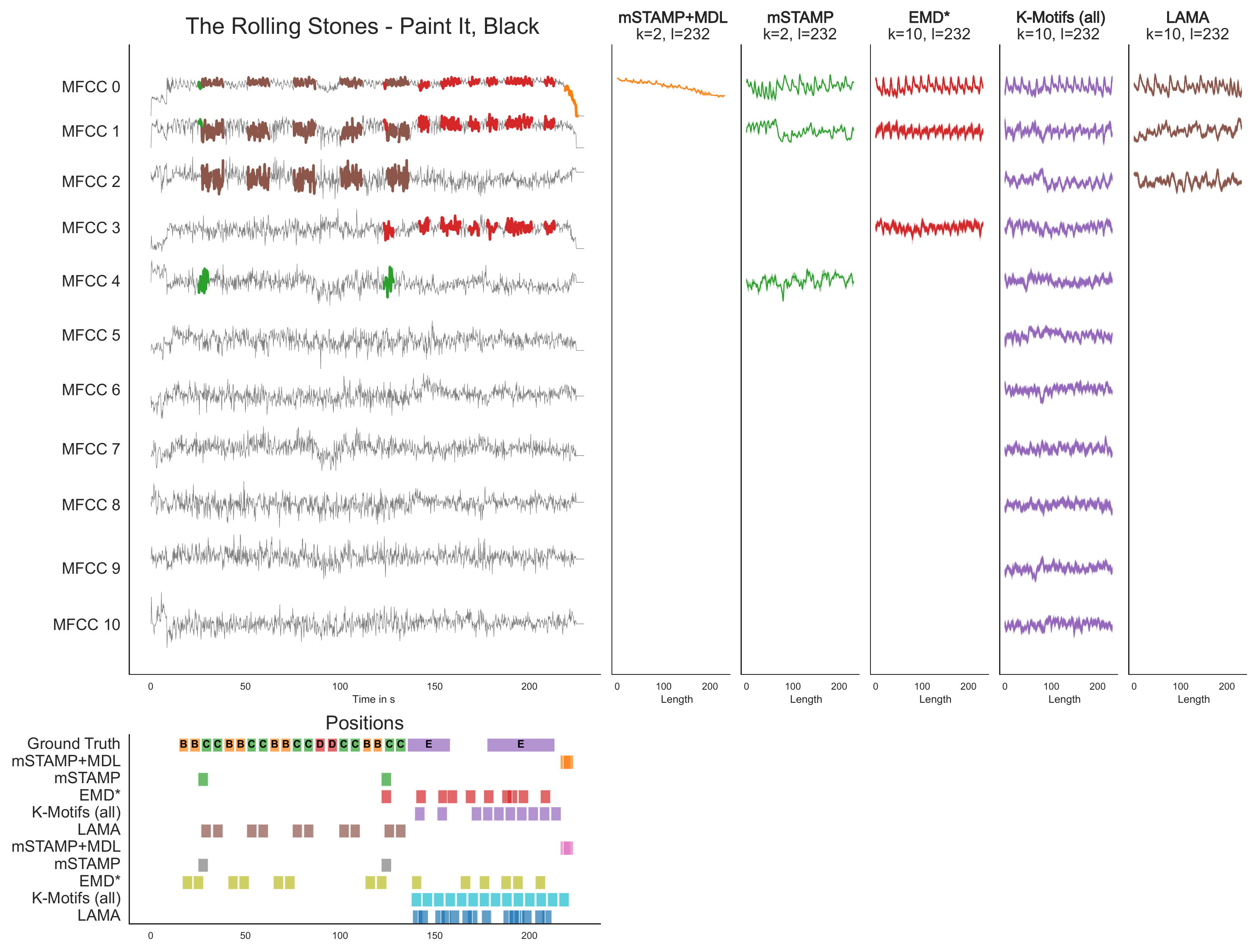}
    \Description[Paint it Black]{Leitmotifs found in Paint it Black Case Study}
	\caption{
	TOP-2 leitmotif discovery. The song \emph{Paint it Black by the Rolling Stones}. The dataset has multiple characteristic sections. We focus on the two largest leitmotifs (in terms of occurrences). One leitmotif has $10$ occurrences and is in the C verse. The second one is the humming of Mick Jagger and Keith Richards in the E section with $14$ occurrences. For the sake of brevity, we display one leitmotifs by method to the right and the positions of both leitmotifs to the bottom. (Image best viewed in color)
	\label{fig:rolling-stones}
	}
\end{figure}

In this experiment, we showcase leitmotifs being discovered in musical scores. Rolling Stones' \emph{Paint it Black}, released in 1966 on the Aftermath album, exemplifies the band's rebellious image. The song is notable for its distinctive use of the sitar, an instrument popularized by the Beatles. The sitar, featured prominently throughout the song, adds a unique Middle Eastern flair. The verses of the song breakdown in the verse structure:

\begin{verbatim}
    B1 – Verse One: I see a red door ...
    C1 – Verse One: I see the girls walk by ...
    B2 – Verse Two
    C2 – Verse Two
    B3 – Verse Three
    C3 – Verse Three
    D1 – Verse Four: No more will ...
    C4 – Verse Four
    B1 – Verse Five
    C1 – Verse Five
    E - Humming
    F - Outro
\end{verbatim}

Instruments are all played in the same style for B and C verses. Overall, the B verse is repeated $4$ times, and the C verse is repeated $5$ times. The B verse contains the iconic \emph{"I see a red door and ..."}. B and C verses break down into two repeats with the same rhythmically. For preprocessing, we extracted the first $7$ MFCCs from the song, equal to 7 channels with 9744 time stamps. The first leitmotif is equivalent to halve of a C verse, which is split at \emph{"I see the girls walk by ..."} and \emph{"I have to turn my head ..."}. It is $5s$ long and repeats twice in each, for a total of $10$ times. The second one is the humming of Mick Jagger and Keith Richards in the E section with $14$ occurrences. 

Figure~\ref{fig:rolling-stones} shows the results. Using LAMA, both leitmotifs in sections C and E were identified. The motif set spans the first three MFCC channels. The second motif set found by LAMA overlaps with the humming section, and is located MFCC channels $0, 1$ and $5$. The B verse was not identified by LAMA. This use case showcases the core concept of LAMA to identify channels by leitmotif. The other competitors use the same channels independent of the leitmotif. mSTAMP, with the number of dimensions as input, identifies two repeats. The B section is found by EMD*, but blurred by additional humming. The remaining motifs found by the other approaches seem to cluster around the humming section E. mSTAMP+MDL struggles to find a meaningful motif. It identified the fade out of the end in the loudness channel MFCC 0.

\paragraph{\textbf{Soundtrack - Imperial March}}

\begin{figure}[t]
    \centering
	\includegraphics[width=1.0\columnwidth]{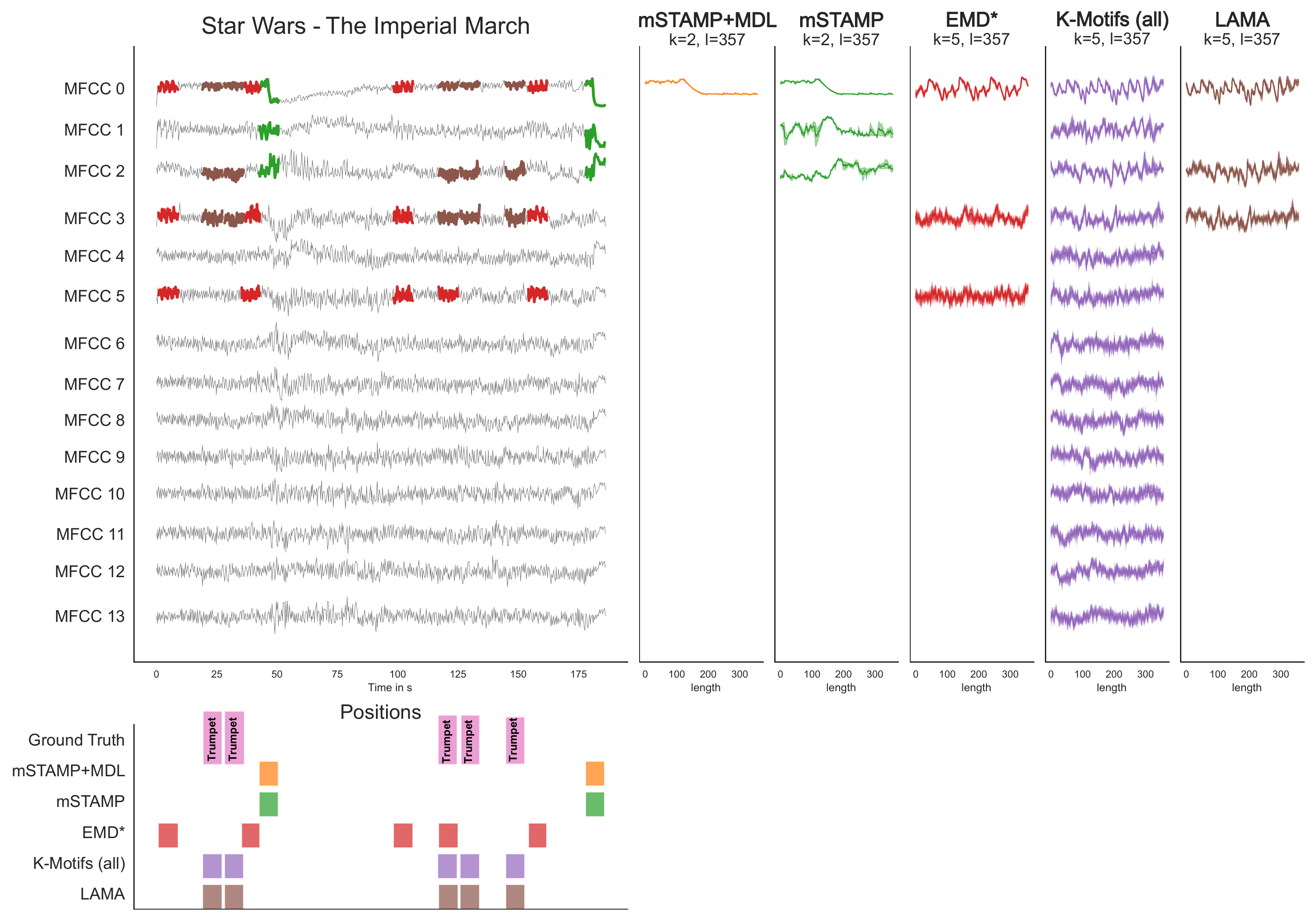}
    \Description[Imperial March]{Leitmotifs found in Imperial March Case Study}
	\caption{
	The Star Wars soundtrack \emph{Imperial March}. The dataset has a characteristic leitmotif played in the presence of Darth Vader. It repeats $5$ times with the trumpets playing for roughly $8.3$ each. (Image best viewed in color)
	\label{fig:star-wars}
	}
\end{figure}

Leitmotifs occur frequently in soundtracks to highlight the appearances of characters or landscapes. John Williams frequently uses leitmotifs in the Star Wars soundtrack. One motif is associated with Darth Vader's appearance, while another with the Death Star or the concept of the Force. The song \emph{Imperial March} features a characteristic leitmotif played by trumpets, and played in the presence of Darth Vader. It has a duration of $8.3$ and repeats $5$ times.
Figure~\ref{fig:star-wars} shows the results. Using Lama, we are able to identify the trumpets playing using $3$ channels $0$, $2$ and $3$. EMD* covers only sub-sections of this leitmotif. It covers sections, where the music increases in volume, too. The best competitor is K-Motifs using all dimensions, finding the leitmotif, too. mSTAMP identifies sections of loudness decrease (fade-out).

\subsection{Scalability}\label{sec:scalability}

\begin{figure}[t]
    \centering
	\includegraphics[width=1.0\columnwidth]{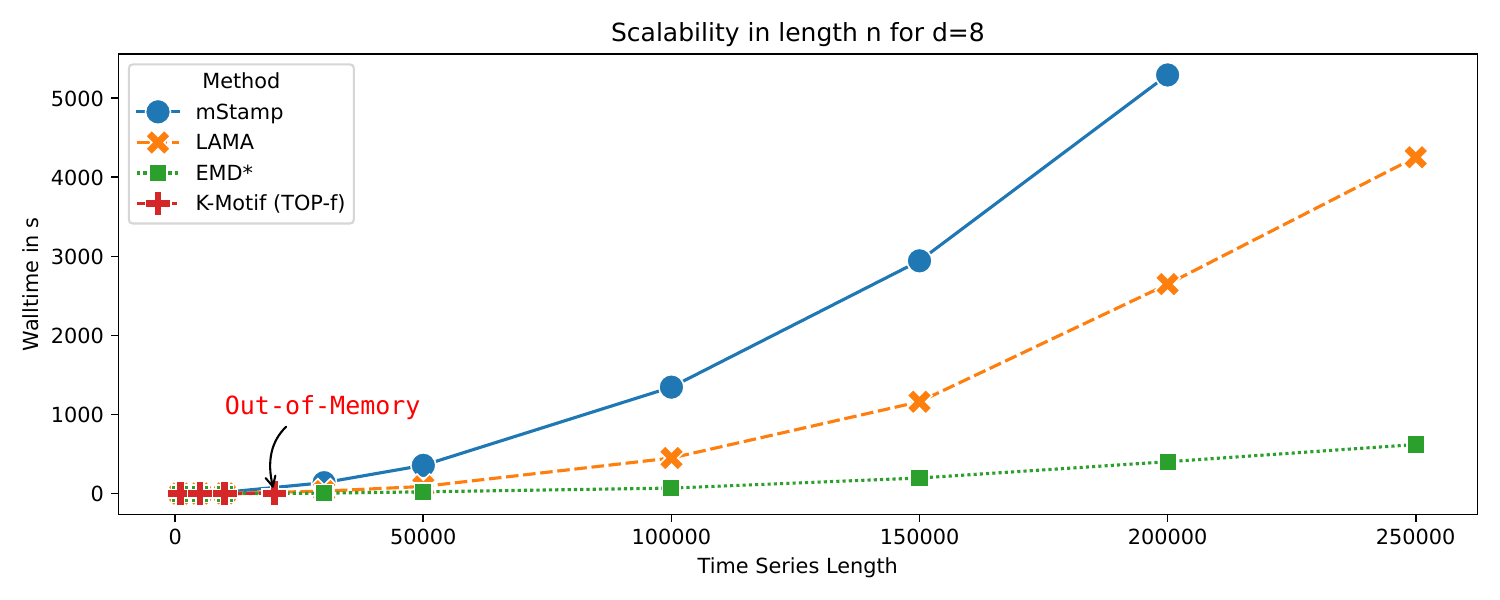}
    \Description[Scalability]{Scalability in length n of the TS}
	\caption{
    Scalability in the length $n$ of the TS.\label{fig:scalability}
	}
\end{figure}

In these experiments, we systematically assessed the impact of increasing TS lengths on the penguin dataset~\cite{zhu2017matrix}, analyzing up to 250k time stamps. The benchmarking was conducted on a laptop equipped with an Apple M1 CPU and 16GB of RAM. Storing a distance matrix of 250k entries in main memory is impractical, requiring approximately $1.8$ terabytes of space ($(250k)^2 \times 8 \times 32$), while our sparse matrix implementation for LAMA (Section~\ref{sec:scalable-motif-sets}) necessitates roughly $12$ gigabytes of RAM, a significant reduction.

Note that mSTAMP, K-Motif, and LAMA have a worst-case time complexity of $\mathcal{O}(n^2 \times d \times L_{max})$, pertaining to the computation of pairwise distances among candidates. EMD* has complexity $\mathcal{O}(n^2 \times L_{max})$ by applying PCA to derive a one dimensional series first, i.e. $d=1$. All implementations are Python-based. For mSTAMP we use the stumpy library, we re-implemented EMD* and K-Motifs.

Figure~\ref{fig:scalability} shows the results of our experiment. All algorithms scale quadratically in $n$. As expected from the complexity analysis, all method exhibit quadratic complexity the TS length, albeit with varying linear factors.
LAMA performs faster than mSTAMP, yet is slower than EMD*, which is due to EMD*'s aggressive PCA dimensionality reduction technique which, however, also severely impacts its pattern extraction quality. K-Motif did not scale beyond 30k, due to its quadratic memory requirements in length $n$.

\subsection{Leitmotifs for Summarization}\label{sec:streaming}

\begin{figure}[t]
    \centering
	\includegraphics[width=1.0\columnwidth]{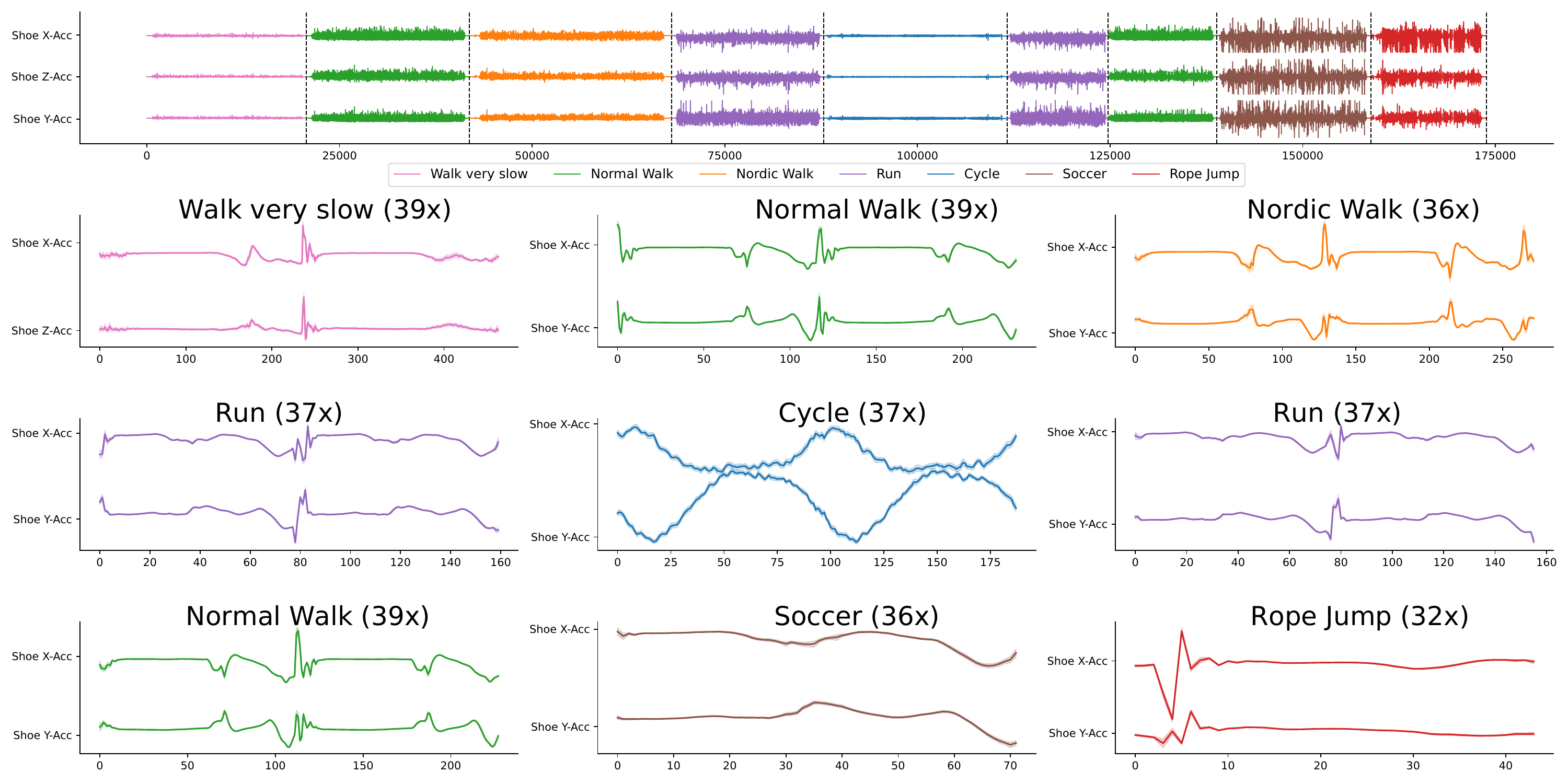}
    \Description[Segmentation]{Leitmotifs in segments}
	\caption{
    The $9$ Leitmotifs found (bottom) by LAMA in a physical outdoor activity by a subject performing $9$ activities (top). First, we performed segmentation, then we ran LAMA on each segment.
	\label{fig:segmentation}
	}
\end{figure}

Leitmotif mining on large TS, such as those with over $150k$ points, can become prohibitive, taking up to $0.5$ hours. However, many TS can first be segmented into semantically disjoint sequences, allowing leitmotif discovery to be performed on these segments. Figure~\ref{fig:segmentation} (top) shows a TS of a subject performing nine outdoor activities for $29$ minutes, spanning 174k data points, sampled at 100Hz, and three channels~\cite{lamprinos2012physical}. If segments are not given, we may apply a segmentation algorithm, such as ClaSS~\cite{ermshaus2023raising}, to derive disjoint segments, which takes ~$70-90s$. As the data is quasi-periodic, we used the dominant Fourier frequency to determine the leitmotif length in each segment. Figure~\ref{fig:segmentation} (bottom) displays the nine leitmotifs and corresponding two channels found for each segment. Notably, \emph{Normal Walk} and \emph{Run} activities are repeated twice, as indicated by the high similarity of the found leitmotifs. Running LAMA on each of the nine segments took a total of $92$s (1900 ops/sec), compared to $739$s (235 ops/sec) when running LAMA on the entire TS. This represents an 8-fold speedup. The approach is faster than real-time ($29$min, or 100Hz). Moreover, it demonstrates that we can summarize a physical activities by their leitmotifs.

\section{Related Work}\label{sec:related–work}

Research in multidimensional motif discovery includes algorithms that find motifs across all dimensions and within subsets (sub-dimensional). Reproducibility is challenging due to the lack of datasets with annotated ground truth, partially addressed by synthetic datasets~\cite{poccia2021smm}. Fine-tuning parameters often relies on trial and error. We present the first systematic benchmark with annotated ground truth from real use cases.
K-Motifs~\cite{lonardi2002finding} are defined as all subsequences within distance radius $r \in \mathbb{R}$ to a centre subsequence of length $l \in \mathbb{N}^{+}$ of the TS. We use K-Motifs as a baseline. 
\emph{mSTAMP}~\cite{yeh2017matrix} extends the matrix profile~\cite{yeh2016matrix} for sub-dimensional pair motif discovery. To avoid enumerating all subsets of dimensions, the algorithm computes 1-NN distances in each dimension separately, then sorts dimensions in ascending order of 1-NN distances. From all $k$ separately computed profiles, a $k$-dimensional cumulative profile is computed along the dimension axis. The optimal number of dimensions is identified using the minimal description length (MDL). 
In~\cite{gao2019discovering}, the variable-length pair motif discovery algorithm \emph{CHIME} is introduced. It transforms subsequences of minimal length $l$ using SAX~\cite{lin2007experiencing} along each dimension, converting the TS into words. These words are merged into larger patterns, and sub-dimensional co-occurring words are correlated across dimensions. Pairwise Euclidean distances between candidates sharing the same word are computed and ranked. CHIME produced runtime errors and only finds pairs, so it was not included in the experiments.
In~\cite{tanaka2005discovery}, the \emph{EMD} method for motif set discovery is introduced. PCA is used to derive a univariate series, and motif sets are found by searching for subsequences within a distance radius $r$ (similar to K-Motifs~\cite{lonardi2002finding}). The TS is SAX transformed to prune the search space, and frequent words are used to find candidate subsequences. Exact Euclidean distance computations within radius $r$ are then performed. MDL identifies the most relevant patterns. No code is available, so we re-implemented relevant parts to use EMD* in our experiments.
In~\cite{minnen2007detecting}, a sub-dimensional method is introduced. The TS is transformed into SAX words, followed by random projections to identify frequent pair motifs. Euclidean distances of corresponding subsequences are computed along each dimension. Relevant dimensions are selected by sorting distances and applying a threshold. The top motif is extended to a set motif by searching for all subsequences within a threshold, similar to VALMOD~\cite{linardi2018valmod}.
For univariate series, VALMOD proofed to be inferior to $k$-Motiflets~\cite{schafer2022motiflets} and K-Motifs~\cite{lonardi2002finding}, and we did not further consider it.
In~\cite{vahdatpour2009toward}, an algorithm for discovering non-synchronous motifs in activity data is presented. It uses random projections of SAX words along each dimension and agglomerative clustering on a coincidence graph, connecting motifs with a lag below a hyper-parameter $\alpha$. Due to the different setting of asynchronous motifs and the lack of available code, we did not include it.
In~\cite{balasubramanian2016discovering} an approach for sub-dimensional variable-length motif discovery is presented tailored for physiological signal data. The approach uses SAX to derive words, and a grammar induction approach to find frequent variable length motif sets. We have requested the code from the authors, but did not receive a response. As it is tailored to one use case, we did not further consider it.
In~\cite{poccia2021smm}, the salient multi-variate motif (SMM) algorithm is introduced for mining sub-dimensional motifs of variable length using metadata. A salient motif is a $k$-frequent subsequence that is distinct from its neighbors and has a pairwise distance of at most $\epsilon$. Patterns are identified through sub-sampling and clustering, using metadata to extract inter-dimensional relations. The method relies on metadata and has $7$ user parameters.

\balance

\section{Conclusion}

A Leitmotif in a time series manifests as a dominant and recurrent subsequence, imbued with semantic meaning. In this work, we introduce a novel algorithm called LAMA for the sub-dimensional discovery of Leitmotifs in time series, and present a novel benchmark of 14 annotated datasets for finding Leitmotifs. Other than previous work, LAMA jointly optimizes motif discovery and channel selection thereby finds considerably more meaningful patterns (Leitmotifs) in each of the benchmark sets. Our findings demonstrate that summarizing a time series through its leitmotifs, as found by LAMA, offers a profound understanding of underlying patterns. This spans a spectrum of activities, from the distinctive punching motion in boxing to the fundamental steps of a dance routine, the wing flap pattern of a penguin, motions form human activity, or the unique vocalizations of a bird. In future work, we aim to further explore the scalability of LAMA by approximating the distance matrix through the application of locality-sensitive hashing.

\begin{acks}
We would like to express our gratitude to Silvestro Poccia, author of SMM, for providing code and dedicating time to emails and Zoom sessions, and for collaborative debugging.  We extend our gratitude to the reviewers for their insightful comments, which significantly enhanced the quality of this paper.
\end{acks}

\bibliographystyle{ACM-Reference-Format}
\bibliography{motiflets}

\end{document}